\documentclass[5p,11pt,twocolumn]{elsarticle}
\usepackage{times,multirow}
\usepackage{url}
\usepackage{color,graphicx,algorithm,algpseudocode}
\usepackage[hang,loose]{subfigure}
\usepackage{amssymb,balance}
\usepackage{pdfpages}

\usepackage[justification=justified]{caption}
\usepackage{color,bm}
\definecolor{Blue}{rgb}{0,0,1}

\begin{document}

\journal{Pervasive and Mobile Computing}

\includepdf[pages={-}]{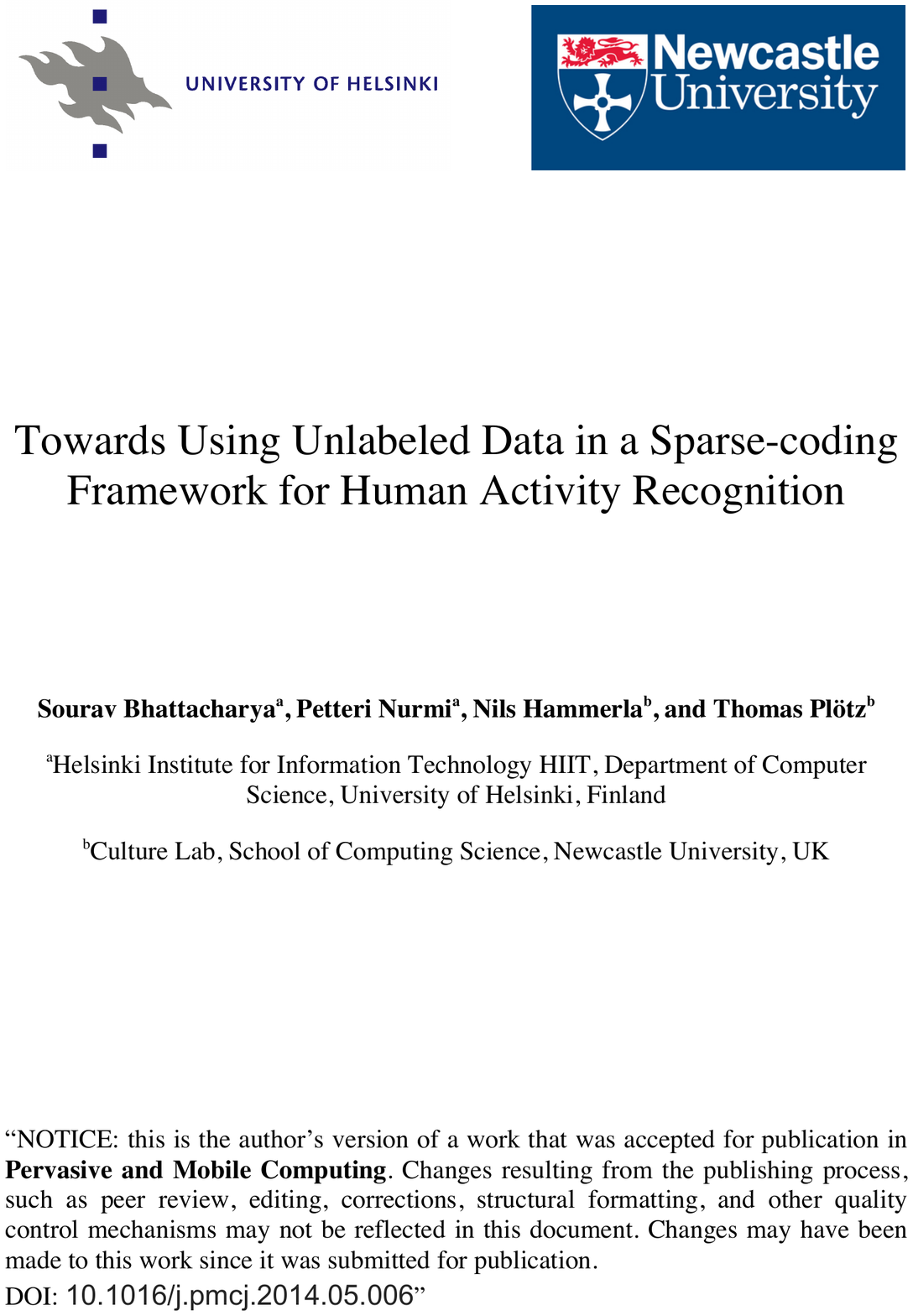}

\begin{frontmatter}

\title{Towards Using Unlabeled Data in a Sparse-coding Framework for Human Activity Recognition}

\author[hiit]{Sourav Bhattacharya}
\author[hiit]{Petteri Nurmi}
\author[ncl]{Nils Hammerla}
\author[ncl]{Thomas Pl{\"o}tz}
\address[hiit]{Helsinki Institute for Information Technology HIIT\\Department of Computer Science, University of Helsinki, Finland}
\address[ncl]{Culture Lab, School of Computing Science, Newcastle University, UK}

\begin{abstract}
We propose a sparse-coding framework for activity recognition in ubiquitous and mobile computing that alleviates two fundamental problems of current supervised learning approaches.
\textit{(i)} It automatically derives a compact, sparse and meaningful feature representation of sensor data that does not rely on prior expert knowledge and generalizes  well across domain boundaries.
\textit{(ii)} It exploits unlabeled sample data for bootstrapping effective activity recognizers, i.e., substantially reduces the amount of ground truth annotation required for model estimation. Such unlabeled data is easy to obtain, e.g., through contemporary smartphones carried by users as they go about their everyday activities.

Based on the self-taught learning paradigm we automatically derive an over-complete set of basis vectors from unlabeled data that captures inherent patterns present within activity data.
Through projecting raw sensor data onto the feature space defined by such over-complete sets of basis vectors effective feature extraction is pursued.
Given these learned feature representations, classification backends are then trained using small amounts of labeled training data.
 
We study the new approach in detail using two datasets which differ in terms of the recognition tasks and sensor modalities. Primarily we focus on a transportation mode analysis task, a popular task in mobile-phone based sensing.
The sparse-coding framework demonstrates better performance than the state-of-the-art in supervised learning approaches.
More importantly, we show the practical potential of the new approach by successfully evaluating its generalization capabilities across both domain and sensor modalities by considering the popular Opportunity dataset. Our feature learning approach outperforms state-of-the-art approaches to analyzing activities of daily living.
\end{abstract}

\begin{keyword}
Activity Recognition, Sparse-coding, Machine Learning, Unsupervised Learning.
\end{keyword}

\end{frontmatter}


\section{Introduction}
\label{sec:introduction}

Activity recognition represents a major research area within mobile and pervasive/ubiquitous computing \cite{atallah09theuse,lane10survey}. 
Prominent examples of domains where activity recognition has been investigated include smart homes~\cite{bao04activity,logan07long,pham09slice}, situated support~\cite{Hoey2011-RSA}, automatic monitoring of mental and physical wellbeing~\cite{ploetz12automatic, consolvo08ubifit, Rabbi2011-PAI}, and general health care~\cite{lester06practical,parkka06activity}. 
Modern smartphones with their advanced sensing capabilities provide a particularly attractive platform for activity recognition as they are carried around by many people while going about their everyday activities.

The vast majority of activity recognition research relies on supervised learning techniques where handcrafted features, e.g., heuristically chosen statistical measures, are extracted from raw sensor recordings, which are then combined with activity labels for effective classifier training.
While this approach is in line with the standard procedures in many application domains of general pattern recognition and machine learning techniques~\cite{bishop07pattern}, it is often too costly or simply not applicable for ubiquitous/pervasive computing applications. 
The reasons for this are twofold.
Firstly, the performance of supervised learning approaches is highly sensitive to the type of feature extraction, where often the optimal set of features varies across different activities \cite{figo10preprocessing,huynh05analyzing,kononen10automatic}. 
Secondly, and more crucially, obtaining reliable ground truth annotation for bootstrapping and training activity recognizers poses a challenge for system developers who target real-world deployments. 
People typically carry their mobile device while going about their everyday activities, thereby not paying much attention to the phone itself in terms of location of the device (in the pocket, in the backpack, etc.) and only sporadically interacting with it (for making a call or explicitly using the device's services for, e.g., information retrieval).
Consequently, active support from users to provide labels for data collected in real-life scenarios cannot be considered feasible for many settings as prompting mobile phone users to annotate their activities while they are pursuing them has its limitations.
Apart from these limitations, privacy and ethical considerations typically render direct observation and annotation impracticable in realistic scenarios. 

Possible alternatives to such direct observation and annotation include: 
\textit{(i)} self-reporting of activities by the users, e.g., using a diary~\cite{huynh08discovery}; 
\textit{(ii)} the use of experience sampling, i.e., prompting the user and asking for the current or previous activity label~\cite{bao04activity, stikic11weakly}; and 
\textit{(iii)} a combination of these methods. 
While such techniques somewhat alleviate the aforementioned problem by providing annotation for at least smaller subsets of unlabeled data, they still remain prone to errors and typically cannot replace expert ground truth annotation. 

Whereas obtaining reliable ground truth annotation is hard to achieve, the collection of, even large amounts of, \textit{unlabeled} sample data is typically straightforward. 
People's smartphones can simply record activity data in an opportunistic way, without requiring the user to follow a certain protocol or scripted activity patterns. 
This is especially attractive since it allows for capturing sensor data while users perform their natural activities without necessarily being conscious about the actual data collection. 

In this paper we introduce a novel framework for activity recognition. 
Our approach mitigates the requirement of large amounts of ground truth annotation by explicitly exploiting \textit{unlabeled} sensor data for bootstrapping our recognition framework. Based on the self-taught learning paradigm~\cite{raina07selftaught}, we develop a sparse-coding framework for unsupervised estimation of sensor data representations with the help of a {\em codebook} of basis vectors (see Section~\ref{sec:system:codebook}). As these representations are learned in an unsupervised manner, our approach also overcomes the need to perform feature-engineering. While the original framework of self-taught learning has been developed mainly for the analysis of non-sequential data, i.e., images and stationary audio signals~\cite{grosse07shift}, we extend the approach towards time-series data such as continuous sensor data streams. We also develop a basis selection method that builds on information theory to generate a codebook of basis vectors that covers characteristic movement patterns in human physical activities. Using {\em activations} of these basis vectors (see Section~\ref{sec:system:fex_and_classifier_training}) we then compute features of the raw sensor data streams, which are the basis for subsequent classifier training. 
The latter requires only relatively small amounts of labeled data, which alleviates the ground truth annotation challenge of mobile computing applications. 

We demonstrate the benefits of our approach using data from two diverse activity recognition tasks, namely transportation mode analysis and classification of activities of daily living (the Opportunity challenge \cite{Roggen10collecting}).
Our experiments demonstrate that the proposed approach provides better results than the state-of-the-art, namely PCA- based feature learning, semi-supervised En-Co-Training, and feature-engineering based (supervised) algorithms, while requiring smaller amounts of training data and not relying on prior domain knowledge for feature crafting. 
Apart from successful generalization across recognition tasks, we also demonstrate 
easy applicability of our proposed framework beyond modality boundaries covering not only accelerometer data but also other commonly available sensors on the mobile platform, such as the gyroscopes, or magnetometers.


\section{Learning From Unlabeled Data}
\label{sec:learningFromUnlabeledData}

The focus of our work is on developing an effective framework that exploits unlabeled data to derive robust activity recognizers for mobile applications.
The key idea is to use vast amounts of easy to record unlabeled sample data for unsupervised {\em feature learning}. 
These features shall cover general characteristics of human movements, which guarantees both robustness and generalizability.

Only very little related work exists that focus on incorporating unlabeled data for training mobile activity recognizers. A notable exception is the work by Amft who explored self-taught learning  in a very preliminary study for activity spotting using on-body motion sensors~\cite{Amft2011-STL}. However, that work does not take into account the properties of the learned codebook which play an important role in the recognition task.

The idea of incorporating unlabeled data and related feature learning techniques into recognizer training is a well researched area in the general machine learning and pattern recognition community. In the following, we will summarize relevant related work from these fields and link them to the mobile and ubiquitous computing domain.

\subsection{Non-supervised Learning Paradigms}
\label{sec:related:paradigms}

A number of general learning paradigms have been developed that focus on deriving statistical models and recognizers by incorporating unlabeled data. 
Although differing in their particular approaches, all related techniques share the objective of alleviating the dependence on a large amount of annotated training data for parameter estimation.

Learning from a combination of labeled and unlabeled datasets is commonly known as \textit{semi-supervised learning}~\cite{chapelle10semisupervised}. The most common approach to semi-supervised learning is {\em generative models}, where the unknown data distribution $p(\bm{x})$ is modeled as a mixture of class conditional distributions $p(\bm{x}| y)$, where $y$ is the (unobserved) class variable. The mixture components are estimated from a large amount of unlabeled and small amount of labeled data by applying the Expectation Maximization (EM) algorithm. The predictive estimate of $p(y|\bm{x})$ is then computed using Bayes' formula. Other approaches to semi-supervised learning include {\em self-training}, {\em co-training}, {\em transductive SVM} (TSVM), {\em graphical models} and {\em multiview learning}.  

Semi-supervised learning techniques have also been applied to activity recognition, e.g., for recognizing locomotion related activities~\cite{guan07activity}, and in smart homes~\cite{stikic08exploring}. 
In order to be effective, semi-supervised learning approaches need to satisfy certain, rather strict assumptions~\cite{chapelle10semisupervised, nigam00text}.
Probably the strongest constraint imposed by these techniques is that they assume that the unlabeled and labeled datasets are drawn from the same distribution, i.e., $\mathcal{D}^u = \mathcal{D}^l$. In other words, the unlabeled dataset has to be collected with strict focus on the set of activities the recognizer shall cover. This limits generalization capability and renders the learning error-prone for real-world settings where the user might perform extraneous activities, or no activity at all~\cite{stiefmeier08wearable}. Our approach provides improved generalization capability by relaxing the equality condition for the distributions of unlabeled and labeled datasets, i.e., $\mathcal{D}^u \ne \mathcal{D}^l$.

An alternative approach to dealing with unlabeled data is \textit{active learning}. Techniques for active learning aim to make the most economic use of annotations by identifying those unlabeled samples that are most uncertain and thus their annotation would provide most information for the training process. Such samples are automatically identified using information theoretic criteria and then manual annotation is requested. Active learning approaches have become very popular in a number of application domains including activity recognition using body-worn sensors~\cite{alemar11using,stikic08exploring}.  Active learning operates on pre-defined sets of features, which stands in contrast to our approach that automatically learns feature representations.
In doing so, active learning becomes sensitive to the particular features that have been extracted, hence limiting its generalizability.

Explicitly focusing on generalizability of recognition frameworks, \textit{transfer learning} techniques have been developed to bridge, e.g., application domains with differing classes or sensing modalities~\cite{Caruana1997-MTL}. In this approach, knowledge acquired in a specific domain can be transferred to another, if a systematic transformation is either provided or learned automatically. Transfer learning has been applied to ubiquitous computing problems, for example, for adapting models learned with data from one smart home to work within another smart home~\cite{hu11cross-domain,kasteren10transferring}, or to adapt activity classifiers learned with data from one user to work with other users~\cite{lane11enabling}. In these approaches the need for annotated training data is not directly reduced but shifted to other domains or modalities, which can be beneficial if such data are easier to obtain. 

As an alternative approach to alleviating the demands of ground truth annotation so-called {\em multi-instance learning} techniques have been developed. These techniques assign labels to sets of instances instead of individual data points~\cite{Amar2001-MIL}. Multi-instance learning has also been applied for activity recognition tasks in ubiquitous computing settings~\cite{stikic11weakly,stikic09activity}. To apply multi-instance learning, labels of the individual instances were considered as hidden variables and a support vector machine was trained to minimize the expected loss of the classification of instances using the labels of the instance sets. Multi-instance learning also operates on a predefined set of features and therefore has limited generalizability.

\subsection{Feature Learning}
\label{sec:related:feature_learning}

Exploiting unlabeled data can also be applied at the feature level to derive a compact and meaningful representation of raw input data.
In fact, feature learning, i.e., unsupervised estimation of suitable data representations, has been actively researched in the machine learning community~\cite{coates11analysis}. The goal of feature learning is to identify and model interesting regularities in the sensor data without being driven by class information. The majority of methods rely on a process similar to generative models but employ efficient, approximative learning algorithms instead of EM~\cite{hinton06fastlearning}.

Data representations for activity recognition in the ubiquitous or mobile computing domain typically correspond to some sort of ``engineered'' feature sets, e.g., statistical values calculated over analysis windows that are extracted using a sliding window procedure~\cite{figo10preprocessing}. Such predefined features often do not generalize across domain boundaries, which requires system developers to optimize their data representation virtually from scratch for every new application domain.
 
Only recently, concepts of feature learning have been successfully applied for activity recognition tasks.
For example, M{\"a}ntyj{\"a}rvi \textit{et al.}~\cite{mantyjarvi01recognizing} compared the use of Principal Component Analysis (PCA) and Independent Component Analysis (ICA) for  extracting features from sensor data. In their approach, either PCA or ICA was applied on raw sensor values. A sliding window was then applied on the transformed data and a Wavelet-based feature extraction method was used in combination with a multilayer perceptron. 

Similarly, Pl{\"o}tz \textit{et al.}\ employed principal component analysis to derive features from tri-axial accelerometer data using a sliding window approach~\cite{plotz11feature}. 
However, instead of applying the PCA on the raw sensor values, they used the empirical cumulative distribution of a data frame to represent the signals before applying PCA \cite{Hammerla2013a}. 
Moreover, they investigated the use of Restricted Boltzmann Machines~\cite{hinton06fastlearning}, to train an autoencoder network for feature learning. 

Minnen \textit{et al.}\ \cite{minnen06discovering} considered activities as sparse motifs in multidimensional time series and proposed an unsupervised algorithm for automatically extracting such motifs from data. 
A related approach was proposed by Frank \textit{et al.}\ \cite{frank10activity} who used time-delay embeddings to extract features from windowed data and fed these features to a subsequent classifier. 

Contrary to the popular Fourier and Wavelet representations, which suffer from non-adaptability to the particular dataset~\cite{hoyer02nonnegative}, we employ a data-adaptive approach of representing accelerometer measurements. The data-adaptive representation is tailored to the statistics of the  data and is directly learned from the recorded measurements. Examples of data-adaptive methods include PCA, ICA and Matrix Factorization. Our approach differs from common data-adaptive methods by employing an {\em over-complete} and sparse feature representation technique. Here, over-completeness indicates that the dimension of the feature space is much higher than the original input data dimension, and sparsity indicates that the majority of the elements in a feature vector is zero.


\section{A Sparse-Coding Framework for Activity\\ Recognition}
\label{sec:system}

We propose a sparse-coding framework for activity recognition that uses a codebook of basis vectors that capture characteristic and latent patterns in the sensor data. As the codebook learning is unsupervised and operates on unlabeled data, our approach effectively reduces the need for annotated ground truth data and overcomes the need to use predefined feature representations, rendering our approach well suited for continuous activity recognition tasks under naturalistic settings. 

\subsection{Method Overview}
\label{sec:system:overview}
\begin{figure*}
\centerline{
	\subfigure[The first phase of sparse-coding based estimation of activity recognizers consists of codebook learning from unlabeled data that results in a codebook of basis vectors that cover characteristic patterns of human movements.]{
	\label{fig:framework:1st}
	\includegraphics[width=0.48\textwidth]{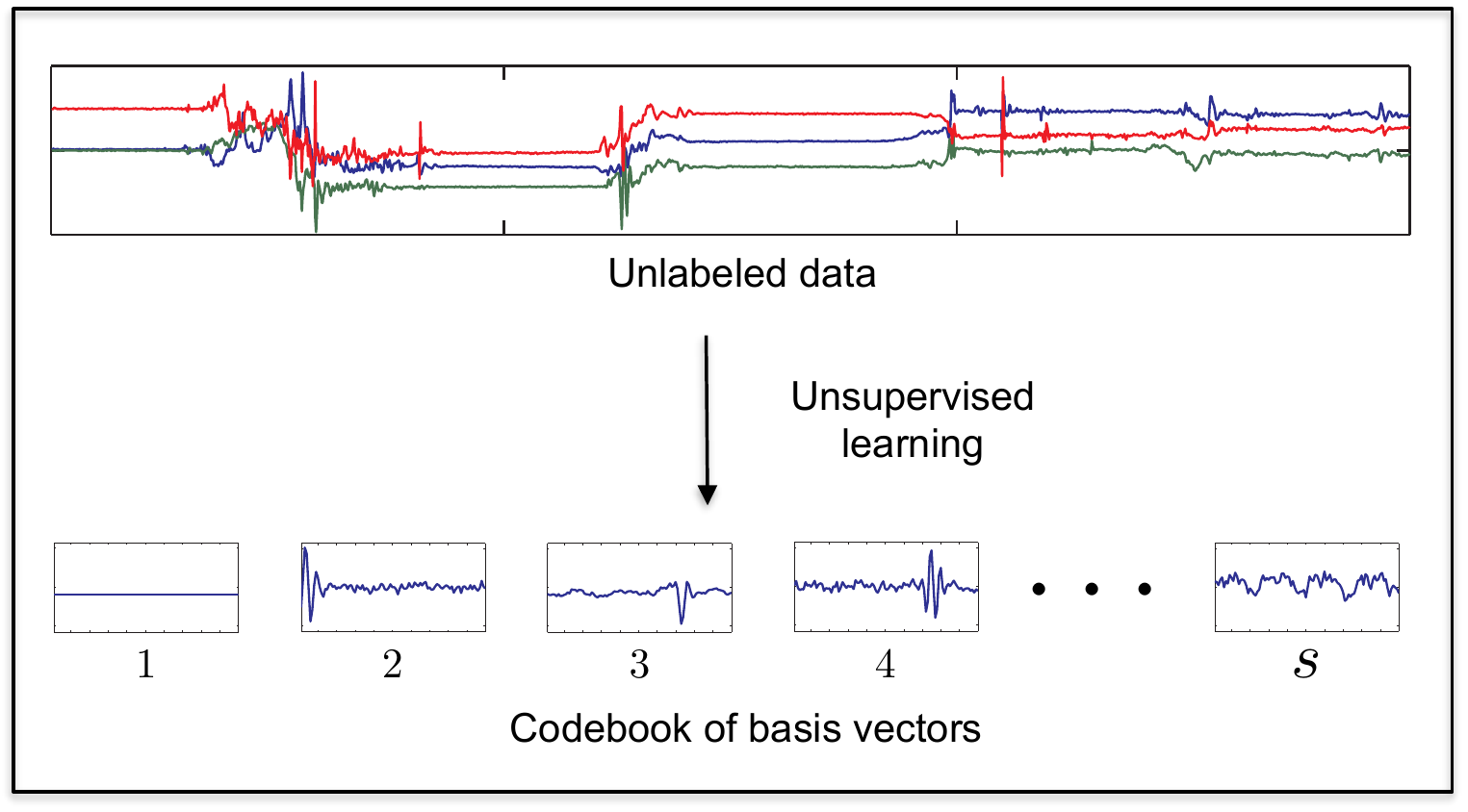}\label{fig:subfig1}}
	\hfill
	\subfigure[The second phase of our modeling approach extracts feature vectors from small amounts of labeled dataset using the codebook of basis vectors extracted in the first phase. Based on these features standard classifier training is performed.]{
	\label{fig:framework:2nd}	
	\includegraphics[width=0.48\linewidth]{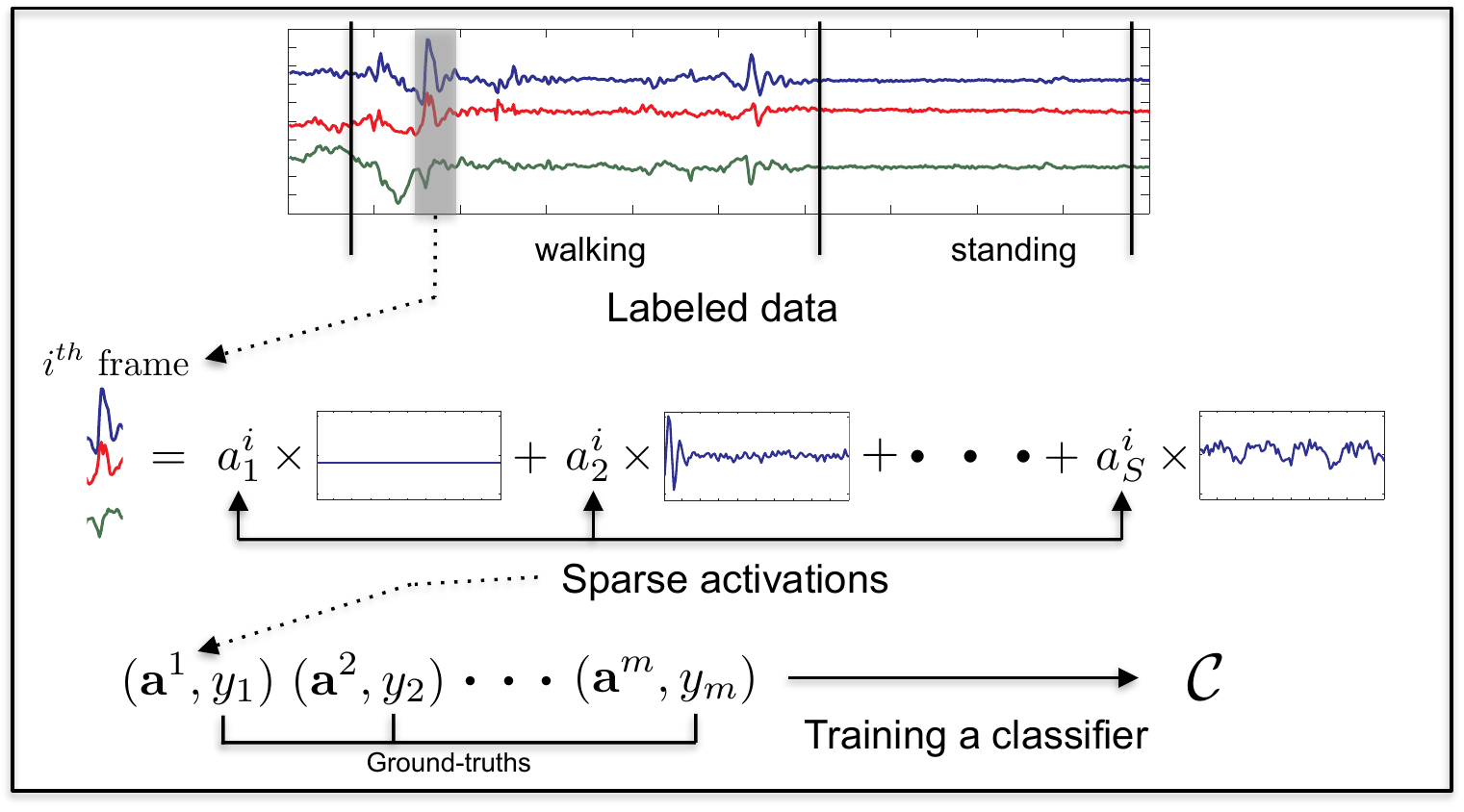}\label{fig:subfig2}}
	}
\caption{Overview of the sparse-coding framework for activity recognition incorporating unlabeled training data.}
\label{fig:framework}
\end{figure*}

Figure~\ref{fig:framework} gives an overview of our approach to learning activity recognizers. We first collect unlabeled data, which in our experiments consists mainly of tri-axial accelerometer measurements (upper part of Figure~\ref{fig:framework:1st}). We then solve an optimization problem (see Section~\ref{sec:system:codebook}) to learn a set of basis vectors --- {\em the codebook} --- that capture characteristic patterns of human movements as they can be observed from the  raw sensor data (lower part of Figure~\ref{fig:framework:1st}). 

Once the codebook has been learned, we use a small set of labeled data to train an activity classifier (Figure~\ref{fig:subfig2}). The features that are used for training the classifier correspond to so-called {\em activations}, which are vectors that enable transferring sensor readings to the feature space spanned by the basis vectors in the codebook. After model training, the activity label for new sensor readings can be determined by transferring the corresponding measurements into the same feature space and applying the learned classifier.

\subsection{Codebook Learning from Unlabeled Data}
\label{sec:system:codebook}

We consider sequential, multidimensional sensor data, which in our experiments correspond to measurements from a tri-axial accelerometer or a gyroscope. We apply a sliding window procedure on the measurements to extract overlapping, fixed length frames. Specifically, we consider measurements of the form $\bm{x}_i \in \mathcal{R}^n$, where $\bm{x}_i$ is a vector containing all measurements within the $i^{th}$ frame and $n$ is the length of the frame, i.e., the unlabeled measurements are represented as the set
\begin{eqnarray}
\bm{X} &=& \{\bm{x}_1, \bm{x}_2, \ldots, \bm{x}_K\},\;  \bm{x}_i \in \mathcal{R}^n.
\end{eqnarray}

In the first step of our approach, we use the unlabeled data $\bm{X}$ to learn a codebook $\bm{\mathcal{B}}$ that captures latent and characteristic patterns in the sensor measurements. The codebook consists of $S$ basis vectors $\{\bm{\beta_j}\}_{j=1}^S$, where each basis vector $\bm{\beta_j} \in \mathcal{R}^n$ represents a particular pattern in the data. Once the codebook has been learned, any frame of sensor measurements can be represented as a linear superposition of the basis vectors, i.e.,
\begin{eqnarray}
\bm{x}_i \approx \sum_{j=1}^S a_j^i \bm{\beta}_j,
\end{eqnarray}
where $a_j^i$ is the activation for $j^{th}$ basis vector when representing the measurement vector $\bm{x}_i$; see Figure~\ref{fig:reconstruction} for an illustration.

The task of learning the codebook $\bm{\mathcal{B}}=\{\bm{\beta}_j\}_{j=1}^S$ from unlabeled data $\bm{X}$ can be  formulated as a regularized optimization problem (see, e.g.,~\cite{lee07efficient,olshausen97sparse,raina07selftaught}). Specifically, we obtain the codebook as the optimal solution to the following minimization problem: 
\begin{eqnarray}
\min_{\bm{\mathcal{B}}, \bm{a}}  \sum_{i=1}^{K} || \bm{x}_i - \sum_{j=1}^{S} a_j^i \bm{\beta}_j||^{2}_{2} + \alpha || \bm{a}^i||_{1}\label{eq:optimization}\\
\mbox{{s.t.}} \;\;\;|| \bm{\beta}_{j} ||_{2} \le 1, \forall j \in \{1, \ldots, S\}.\nonumber 
\end{eqnarray}
Equation~\ref{eq:optimization} contains two optimization variables: (i) the codebook $\bm{\mathcal{B}}$; and (ii) the activations $\bm{a} = \{\bm{a}^1,\bm{a}^2,\ldots,\bm{a}^K\}$. 
The regularization parameter $\alpha$ controls the trade-off between reconstruction quality and sparseness of the basis vectors. Smaller values of $\alpha$ lead to the first term, i.e., the quadratic term in Equation~\ref{eq:optimization}, dominating, thereby generating basis vectors whose weighted combination can represent input signals accurately. In contrast, large values (e.g., $\alpha \approx 1$) shift the importance towards the regularization term, thereby encouraging sparse solutions where the activations have small $L_1$-norm, i.e., the input signal is represented using only a few basis vectors. 

The constraint on the norm of each basis vector $\bm{\beta}_j$ is essential to avoid trivial solutions, e.g., very large $\bm{\beta}_j$ and very small activations $\bm{a}^i$~\cite{hoyer02nonnegative}.  Note that Equation \ref{eq:optimization} does not pose any restrictions on the number of basis vectors $S$ that can be learned. In fact, the codebook can be over-complete, i.e., containing more basis vectors than the input data dimension (i.e., $S \gg n$). Over-completeness reduces sensitivity to noise, whereas the application of sparse-coding enables deviating from a purely linear relationship between the input and output, enabling the codebook to capture complex and high-order patterns in the data~\cite{raina07selftaught,olshausen97sparse}. 

The minimization problem specified in Equation~\ref{eq:optimization} is not convex on both $\bm{\mathcal{B}}$ and $\bm{a}$ simultaneously. However, it can easily be divided into two convex sub-problems, which allows for iterative optimization of both $\bm{\mathcal{B}}$ and $\bm{a}$, thereby keeping one variable constant while optimizing the other. Effectively this corresponds to solving a L$_2$-constrained least squares problem while optimizing for $\bm{\mathcal{B}}$ keeping $\bm{a}$ constant, followed by optimizing $\bm{a}$ whilst keeping $\bm{\mathcal{B}}$ constant, i.e., solving an L$_1$-regularized least square problem~\cite{raina07selftaught}. The solution to the optimization problem, specifically in the case of a large dataset and highly over-complete representation, is computationally expensive~\cite{lee07efficient}. Following Lee \textit{et al.}\ \cite{lee07efficient}, we use a fast iterative algorithm to codebook learning. Algorithm~\ref{alg:codebook} summarizes the procedure, where the FeatureSignSearch algorithm (line \ref{fsign}) solves the L$_1$-regularized least square problem (for details see~\cite{lee07efficient}). The codebook is derived by using standard least square optimization (line \ref{lagrange}). The convergence of the algorithm is detected when the drop in the objective function given in Equation~\ref{eq:optimization} is insignificant between two successive iterations.

\begin{algorithm}[!t]
\begin{algorithmic}[1]
\State {\bf Input:} Unlabeled dataset $\bm{X} = \{\bm{x}_i\}_{i=1}^K$
\smallbreak
\State {\bf Output:} Codebook $\bm{\mathcal{B}} = \{\bm{\beta}_j\}_{j=1}^S$
\smallbreak
\State {\bf Algorithm:}
\For{$j \in \{1,\ldots,S\}$} \Comment Initializing basis vectors
	\State $\bm{\beta}_j \sim U(-0.5,0.5)$
	\State $\bm{\beta}_j = $ MeanNormalize$(\bm{\beta}_j)$
	\State $\bm{\beta}_j = $ MakeNormUnity$(\bm{\beta}_j)$
\EndFor
\Repeat
\State $\{Batch_q\}_{q=1}^M$ = Partition($\bm{X}$) \Comment Randomly partition data into $M$ batches
\For{$q \in \{1,\ldots,M\}$}
\State $\bm{a}_{Batch_q} = $ FeatureSignSearch($Batch_q, \bm{\mathcal{B}}$)\label{fsign}
\State $\bm{\mathcal{B}} = $  LeastSquareSolve($Batch_q, \bm{a}_{Batch_q}$)\label{lagrange}
\EndFor
\Until{convergence}
\State {\bf return:} $\bm{\mathcal{B}}$
\end{algorithmic}
\caption{Fast Codebook Learning}
\label{alg:codebook}
\end{algorithm}

\subsubsection*{Codebook Selection}
\label{sec:system:pruning}

When sparse-coding is applied on sequential data streams, the solution to the optimization problem specified by Equation~\ref{eq:optimization} has been shown to produce redundant basis vectors that are structurally similar, but shifted in time~\cite{grosse07shift}. Grosse \textit{et al.}\ have proposed a convolution technique that helps to overcome redundancy by allowing the basis vectors $\bm{\beta}_j$ to be used at all possible time shifts within the signal ${\bm x_i}$. Specifically, in this approach the optimization equation is modified into the following form:
\begin{eqnarray}
\min_{\bm{\mathcal{B}}, \bm{a}}  \sum_{i=1}^{K} || \bm{x}_i - \sum_{j=1}^{S}  \bm{\beta}_j * \bm{a}_j^i||^{2}_{2} + \alpha || \bm{a}^i||_{1}\label{eq:optimizationConv}\\
\mbox{{subject to}} \;\;\;|| \bm{\beta}_{j} ||_{2} \le c, \forall j \in \{1, \ldots, S\},\nonumber 
\end{eqnarray}
where $\bm{x}_i \in \mathcal{R}^n$ and $\bm{\beta}_j \in \mathcal{R}^p$ with $p \le n$. The activations are now $n-p+1$ dimensional vectors, i.e., $\bm{a}_j^i \in \mathcal{R}^{n-p+1}$, and the measurements are represented using a convolution of activations and basis vectors, i.e., $\bm{x}_i = \bm{\beta}_i * \bm{a}_j^i$. However, this approach is computationally intensive, rendering it unsuitable to mobile devices. Instead of modifying the optimization equation itself, we have developed a basis vector selection technique based on an information theoretic criterion. The selection procedure reduces redundancy by removing specific basis vectors that are structurally similar. 

\begin{figure}[!t]
\begin{center}
\includegraphics[width=\linewidth]{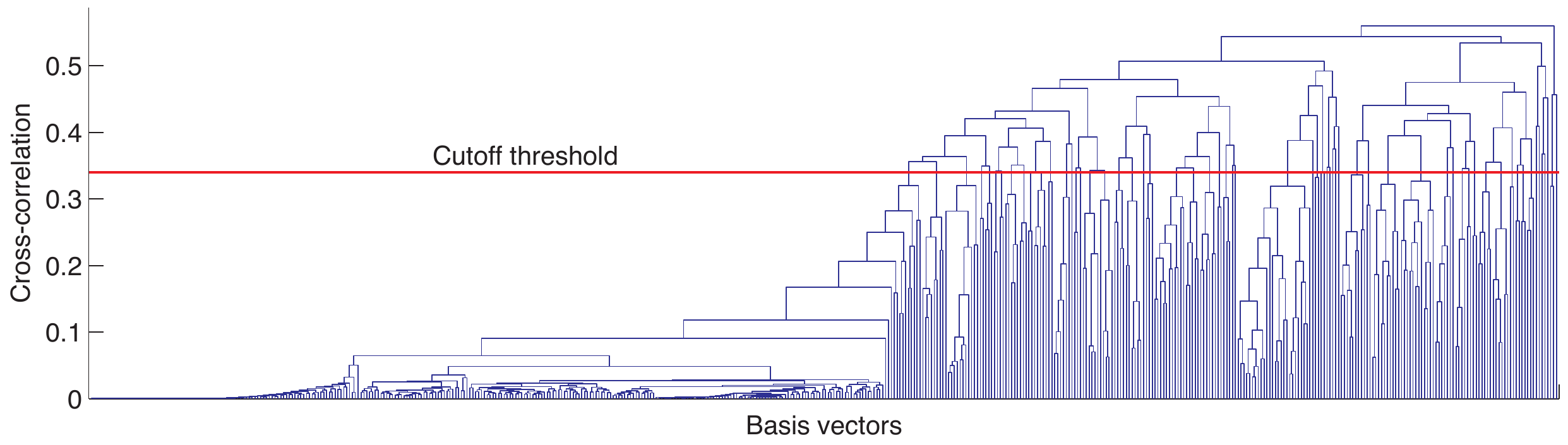}
\caption{Dendrogram showing the hierarchical relationship, with respect to cross-correlation, present within a codebook of $512$ basis vectors. The plot also indicates the cutoff threshold used to generate $52$ clusters.}
\label{fig:dendrogram}
\end{center}
\end{figure}

In the first step of our codebook selection technique, we employ a hierarchical clustering of the basis vectors. More specifically, we use the complete linkage clustering algorithm~\cite{berkhin06survey} with maximal cross-correlation as the similarity measure between two basis vectors:
\begin{eqnarray}
& sim(\bm{\beta}, \bm{\beta}') = \\\nonumber
& \max\sum\limits_{\tau=\max(1,t-n+1)}^{\min(n,t)}\bm{\beta}(\tau) \bm{\beta'}(n+\tau-t).
\end{eqnarray}
The clustering returns a hierarchical representation of the similarity relationships between the basis vectors. From this hierarchy, we then select a subset of basis vectors that contains most of the information. In order to do so, we first apply an adaptive cutoff threshold on the hierarchy to divide the basis vectors into $\lceil S / 10 \rceil$ clusters. For illustration, Figure~\ref{fig:dendrogram} shows the dendrogram plot of the hierarchical relationships found within a codebook of $512$ basis vectors. The red line in the figure indicates the cutoff threshold ($0.34$) used to divide the basis vectors into $52$ clusters. Next, we remove from each cluster those basis vectors that are not sufficiently informative. Specifically, we order the basis vectors within a cluster by their empirical entropy\footnote{To calculate the empirical entropy, we construct a histogram of the basis vector values. The empirical entropy is then computed by $-\sum_q p_q \cdot\log p_q$, where $p_q$ is the probability of the $q^{th}$ histogram bin.} and discard the lowest 10-percentile of vectors. The basis vectors that remain after this step constitute the final codebook $\bm{\mathcal{B}}^*$ used by our approach.

\subsection{Feature Representations and Classifier Training}
\label{sec:system:fex_and_classifier_training}

Once the codebook has been learned, we use a small set of labeled data to train a classifier that can be used to determine the appropriate activity label for new sensor readings. Let $\bm{X'} = \{ \bm{x}_1',\ldots,\bm{x}_M'\}$ denote the set of measurement frames for which ground truth labels are available and let $y = \left( y_1, \ldots, y_M\right)$ denote the corresponding activity labels. To train the classifier, we first map the measurements in the labeled dataset to the feature space spanned by the basis vectors. Specifically, we need to derive the optimal activation vector $\bm{\widehat{a}}^i$ for the measurement $\bm{x}_i'$, which corresponds to solving the following optimization equation:
\begin{eqnarray}
\bm{\widehat{a}}^i = \arg \min_{\bm{a}_i} || \bm{x}_i' - \sum_{j=1}^S a_j^i \bm{\beta}_j||_ 2^2 + \alpha || \bm{a}^i ||_1\label{feature}.
\end{eqnarray}
Once the activation vectors $\bm{\widehat{a}}^i$ have been calculated, a supervised classifier is learned using the activation vectors as features and the labels $y_i$ as the class information, i.e., the training data consists of tuples $(\bm{\widehat{a}^i},y_i)$. The classifier is learned using standard supervised learning techniques. In our experiments we consider decision trees, nearest-neighbor, and support vector machines (SVM) as the classifiers; however, our approach is generic and any other classification technique can be used.

To determine the activity label for a new measurement frame $\bm{x}_q$, we first map the measurement onto the feature space specified by the basis vectors in the codebook, i.e., we use Equation~\ref{feature} to obtain the activation vector $\bm{\widehat{a}}^q$ for $\bm{x}_q$. The current activity label can then be determined by giving the activation vector $\bm{\widehat{a}}^q$ as input to the previously trained classifier.

The codebook selection procedure based on hierarchical clustering also helps to improve the running time of above optimization problem while extracting feature vectors and therefore suits well for mobile platforms.
The overall procedure of our sparse-coding based framework for activity recognition is summarized in Algorithm~\ref{alg:activity}.

\begin{algorithm}[!t]
\begin{algorithmic}[1]
\State {\bf Input:} Unlabeled dataset $\bm{X} = \{\bm{x}_i\}_{i=1}^K$ and \\
\hspace{11 mm}Labeled dataset $\bm{X'} = \{(\bm{x}_i', y_i)\}_{i=1}^M$. 
\smallbreak
\State {\bf Output:} Classifier $\mathcal{C}$
\smallbreak
\State {\bf Algorithm:}
\State $\bm{\mathcal{B}} = $ Fast Codebook Learning($\bm{X}$) \Comment Learning a codebook from unlabeled data using Algorithm~\ref{alg:codebook}
\State  Identify clusters $\{\mathcal{K}_i\}_{i=1}^C$ within learned codebook $\bm{\mathcal{B}}$ based on structural similarities.
\State $\bm{\mathcal{B}}^* = \emptyset$ \Comment{Initialization of optimized codebook}
\For{ $j \in \{1,\ldots,C\}$} 
	\State{$\bm{\mathcal{B}}^* = \bm{\mathcal{B}}^* \cup $ Select($\mathcal{K}_j$)} \Comment{selection of most informative basis vectors from a cluster}
\EndFor
\State $\mathcal{F} = \emptyset$ \Comment{Initialization of feature set}
\For{$i \in \{1,\ldots,M\}$}
	\State $\bm{\widehat{a}}^i = \arg \min_{\bm{a}_i} || \bm{x'}_i - \sum_{j=1}^{S^*} a_j^i \bm{\beta}_j||_ 2^2 + \alpha || \bm{a}^i ||_1$ \Comment{Here, $\bm{\beta}_j \in \bm{\mathcal{B}}^*,\; \forall j$ and $S^* = |\bm{\mathcal{B}}^*|$}
	\State $\mathcal{F} = \mathcal{F} \cup (\widehat{\bm{a}}_i, y_i)$
\EndFor
\State $\mathcal{C}=\mbox{ClassifierTrain}(\mathcal{F})$
\State {\bf return:} $\mathcal{C}$
\end{algorithmic}
\caption{Sparse-code Based Activity Recognition}
\label{alg:activity}
\end{algorithm}


\section{Case Study: Transportation Mode Analysis}
\label{sec:evaluation}

In order to study the effectiveness of the proposed sparse-coding framework for activity recognition, we conducted an extensive case study on transportation mode analysis.
We utilized smartphones and their onboard sensing capabilities (tri-axial accelerometers) as mobile recording platform to capture people's movement patterns and then use our new activity recognition method to detect the transportation modes of the participants in their everyday life, e.g., walking, taking the metro and riding the bus.

Knowledge of transportation mode has relevance to numerous fields, including human mobility modeling~\cite{lazer09computational}, inferring transportation routines and predicting future movements~\cite{liao07learning}, urban planning~\cite{zheng11urban}, and emergency response, to name but a few~\cite{soper12human}. 
It is considered as a representative example of mobile computing applications \cite{lane10survey}.
Gathering accurate annotations for transportation mode detection is difficult as the activities take place in everyday situations where environmental factors, such as crowding, can rapidly influence a person's behavior. 
Transportation activities are also often interleaved and difficult to distinguish (e.g., a person walking in a moving bus on a bumpy road).
Furthermore, people often interact with their phones while moving, which adds another level of interference and noise to the recorded signals.

The state-of-the-art in transportation mode detection largely corresponds to feature-engineering based approaches~\cite{brezmes09activity,reddy10using,wang10accelerometer,hemminki13accelerometer}, which we will use as a baseline for our evaluation. 

\subsection{Dataset}
\label{ssec:tmd}

For the case study we have collected a dataset that consists of 
approximately $6$ hours of 
consecutive accelerometer recordings.
Three participants, graduate students in Computer Science who had prior experience in using touch screen phones, carried three Samsung Galaxy S II each while going about everyday life activities. The participants were asked to travel between a predefined set of places with a specific means of transportation.  The data collected by each participant included still, walking and traveling by tram, metro, bus and train. 
The phones were placed at three different locations: 
\textit{(i)} jacket's pocket; 
\textit{(ii)} pants' pocket; and 
\textit{(iii)} backpack. 

Accelerometer data were recorded with a sampling frequency of $100$ Hz. 
For ground truth annotation participants were given another mobile phone that was synchronized with the recording devices and provided a simple annotation GUI.
The dataset is summarized in Table~\ref{summary}. 

\begin{table}
\centering
\begin{tabular}{c||ccc||c}
\multicolumn{1}{l}{\bf User} & {\bf Bag} & {\bf Jacket} & \multicolumn{1}{c}{{\bf Pant}} & {\bf Hour} \\
\hline
\hline
{\bf  1} & $681,913$ & $558,632$ & $682,012$ & $2.1$\\
{\bf  2} & $532,773$ & $535,310$ & $532,354$ & $1.7$\\
{\bf  3} & $613,024$ & $600,471$ & $611,502$ & $1.9$\\
\hline
\hline
{\bf Total} & $1,827,710$ & $1,694,413$ & $1,825,868$ & $5.7$\\
\end{tabular}
\caption{Summary of the dataset used for the case study on transportation mode analysis. The first three columns contain the number of samples recorded from each phone location, and the final column shows the overall duration of the corresponding measurements (in hours).}
\label{summary}
\end{table}%

\subsection{Pre-processing}

Before applying our sparse-coding based activity recognition framework, the recorded raw sensor data have to undergo certain standard pre-processing steps.

\subsubsection*{Orientation Normalization}
Since tri-axial accelerometer readings are sensitive to the orientation of the particular sensor, we consider magnitudes of the recordings, which effectively normalizes the measurements with respect to the phone's spatial orientation.
Formally, this normalization corresponds to aggregating the tri-axial sensor readings 
using the $L_2$-norm, i.e., we consider measurements of the form $d = \sqrt{ d_x^2 + d_y^2 + d_z^2}$ where $d_x, d_y$ and $d_z$ are the different acceleration components at a time instant. 
Magnitude-based normalization corresponds to the state-of-the-art approach for achieving rotation invariance in smartphone-based activity recognition~\cite{reddy10using,wang10accelerometer,lu10jigsaw}. 

\subsubsection*{Frame Extraction}
For continuous sensor data analysis we extract small analysis frames, i.e., windows of consecutive sensor readings, from the continuous sensor data stream. 
We use a sliding window procedure~\cite{plotz11activity} that circumvents the need for explicit segmentation of the sensor data stream, which in itself is a non-trivial problem.
We employ a window size of one second, corresponding to $100$ sensor readings. 
Using a short window length enables near real-time information about the user's current transportation mode and ensures the detection can rapidly adapt to changes in transportation modalities~\cite{reddy10using,lu10jigsaw}. 
Consecutive frames overlap by $50\%$ and the activity label of every frame is then determined using majority voting. For example, in our analysis two successive frames have exactly $50$ contiguous measurements in common and the label of a frame is determined by taking the most frequent ground-truth label of the $100$ measurements present within it. In (rare) cases of a tie, the frame label is determined by selecting randomly among the labels with the highest occurrence frequency.

\subsection{Codebook Learning}
\label{ssec:cblearning}

According to the general idea of our sparse-coding based activity recognition framework, we derive a user-specific codebook of basis vectors from unlabeled frames of accelerometer data (magnitudes) 
by applying the fast codebook learning algorithm as described in the previous section (see Algorithm~\ref{alg:codebook}). 
With a sampling rate of $100$ Hz and a frame length of $1$s, the dimensionality of both input $\bm{x}_{i}$ and the resulting basis vectors $\bm{\beta}_j$ is $100$, i.e., $\bm{x}_{i} \in \mathcal{R}^{100}$ and $\bm{\beta}_{j} \in \mathcal{R}^{100}$.

\begin{figure}[!t]
\centering
\subfigure[Examples of basis vectors learned from accelerometer data.]{
\includegraphics[width=\linewidth]{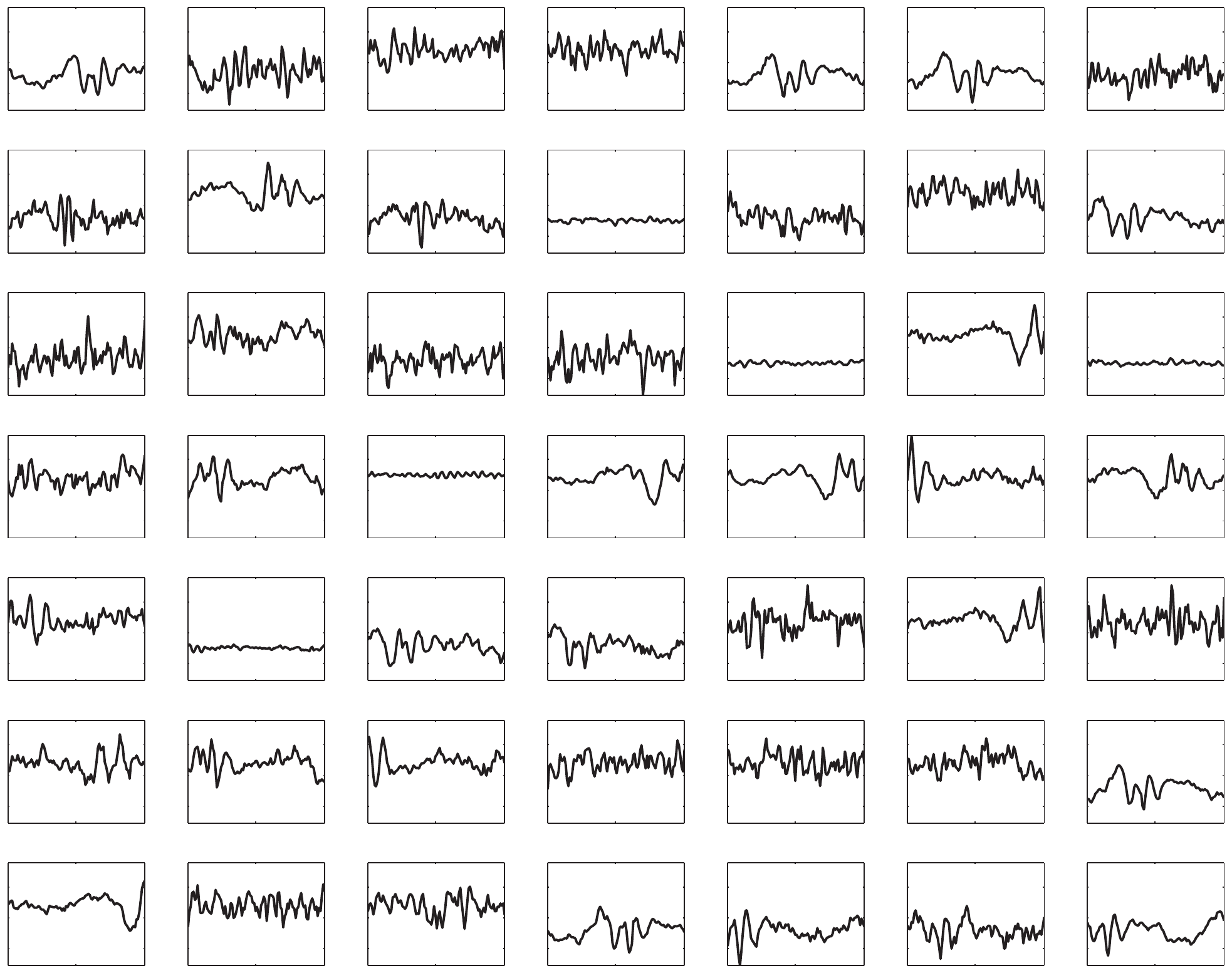}\label{fig:TMbases}}
\subfigure[Examples of basis vectors from one cluster, showing the time-shifting property.]{
\includegraphics[width=\linewidth]{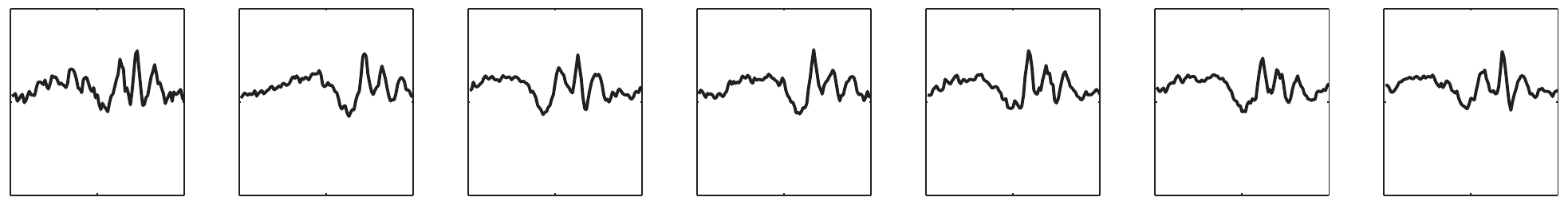}\label{fig:TMbases2}}
\caption{Examples of basis vectors as learned from the transportation mode dataset and example of the time-shifting property observed within a codebook.}
\end{figure}

Figure~\ref{fig:TMbases} illustrates the results of the codebook learning process by means of $49$ exemplary basis vectors as they have been derived from one arbitrarily chosen participant (User 1). 
The shown basis vectors were randomly picked from the generated codebook. 
For illustration purposes Figure~\ref{fig:TMbases2} additionally shows examples of the time-shifted property observed within the learned set of basis vectors.

By analyzing the basis vectors it becomes clear that: 
\textit{(i)} the automatic codebook selection procedure covers a large variability of input signals; and 
\textit{(ii)} 
that basis vectors assigned to the same cluster often are time-shifted variants of each other.

\begin{figure}[!t]
\begin{center}
\subfigure[Reconstruction of a measurement frame using $54$ basis vectors from a codebook containing $512$ basis vectors.]{
	\includegraphics[width=0.7\linewidth]{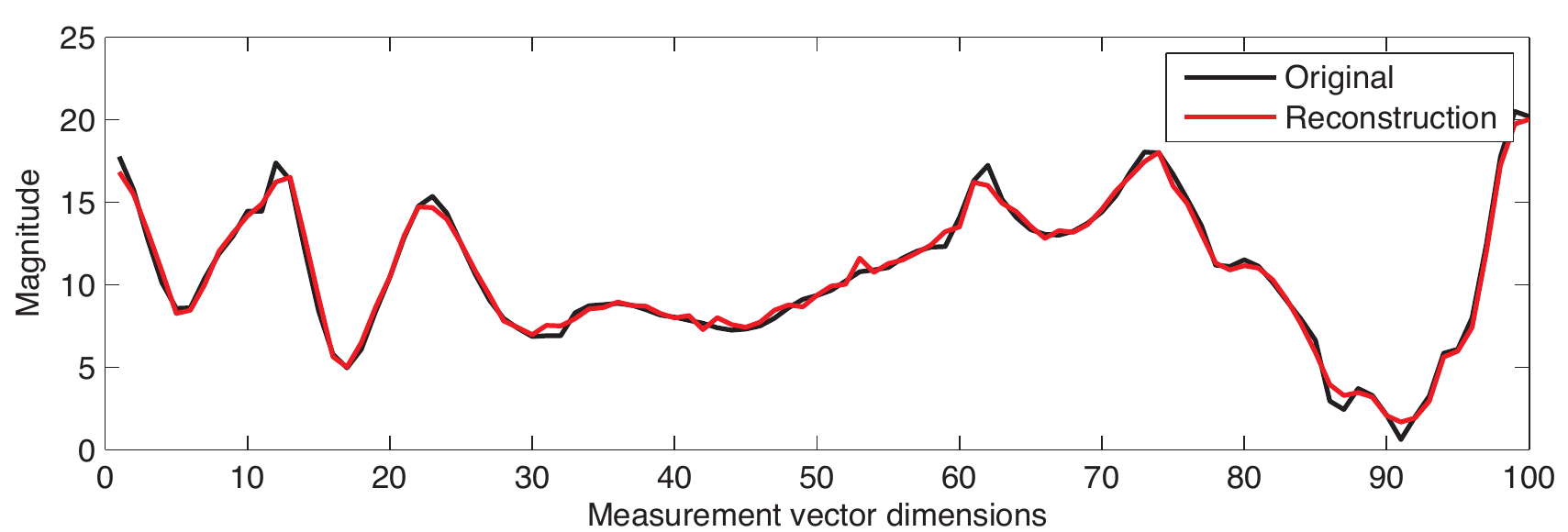}\label{fig:reconstruction}}
\subfigure[Histograms of number of basis vector activations.]{
	\includegraphics[width=\linewidth]{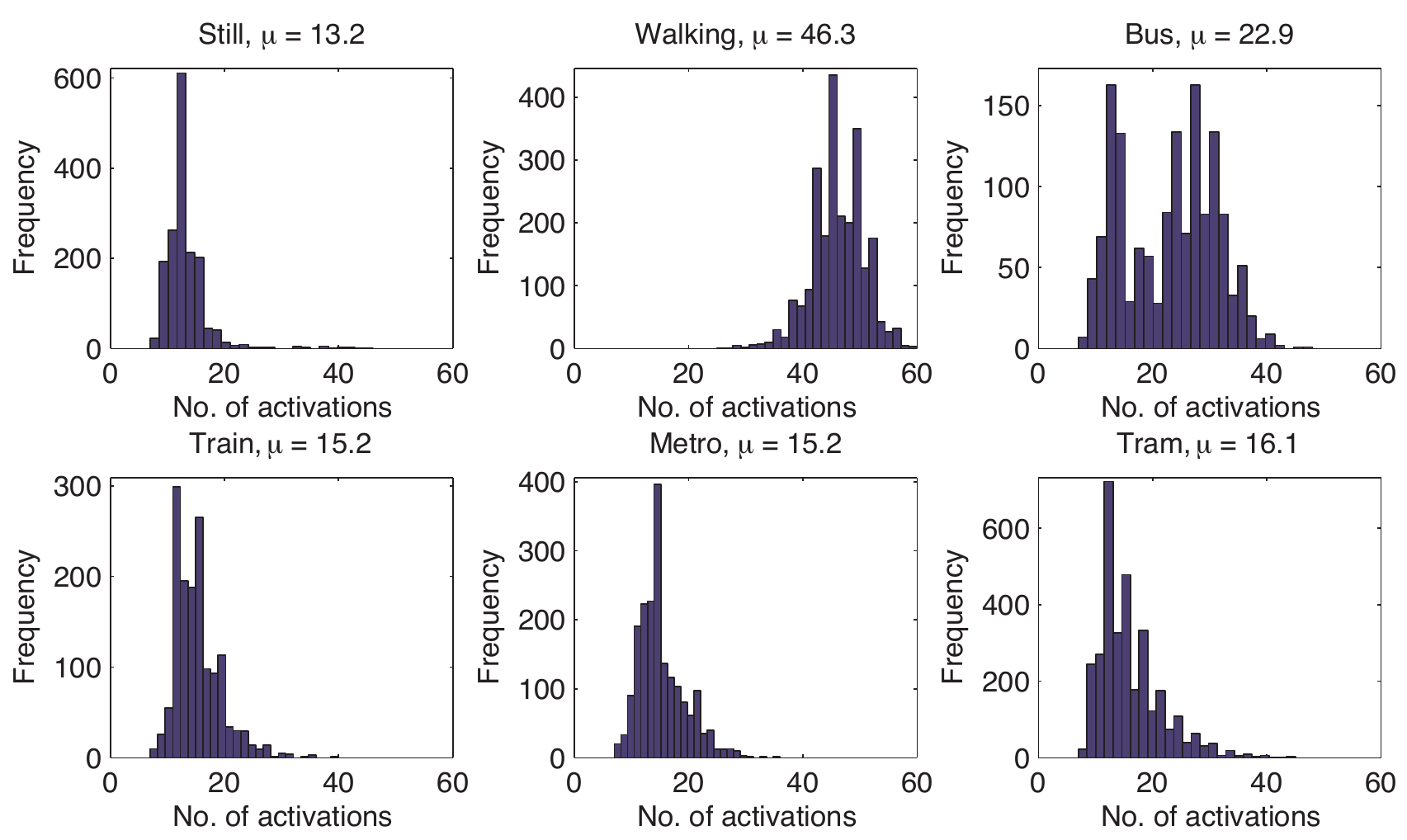}\label{fig:approx}}
\caption{(a) Example of reconstruction of a frame of accelerometer measurements (after normalization). (b) Histograms showing the frequency distributions of the number of basis vectors activated for the reconstruction of accelerometer measurement frames for different transportation modes present in one dataset (User 1). The figure also indicates the average number of basis vector activations per transportation mode.}
\end{center}
\end{figure}

When representing an input vector, the activations of basis vectors 
are sparse, i.e., only a small subset of the over-complete codebook has non-zero weights that originate from different pattern classes.
The sparseness property improves the discrimination capabilities of the framework. For example, Figure~\ref{fig:reconstruction} illustrates the reconstruction of an acceleration measurement frame with $54$ out of $512$ basis vectors present in a codebook. Moreover, Figure~\ref{fig:approx} illustrates the histograms of the number of basis vectors activated to reconstruct the measurement frames, specific to different transportation modes, for the dataset collected by User 1. The figure indicates that a small fraction of the basis vectors from the learned codebook (i.e., $\ll 512$) are activated to accurately reconstruct most of the measurement frames. 

\begin{figure}[!h]
\begin{center}
\subfigure[]{
	\label{fig:recon:distribution}
	\includegraphics[width=0.41\linewidth]{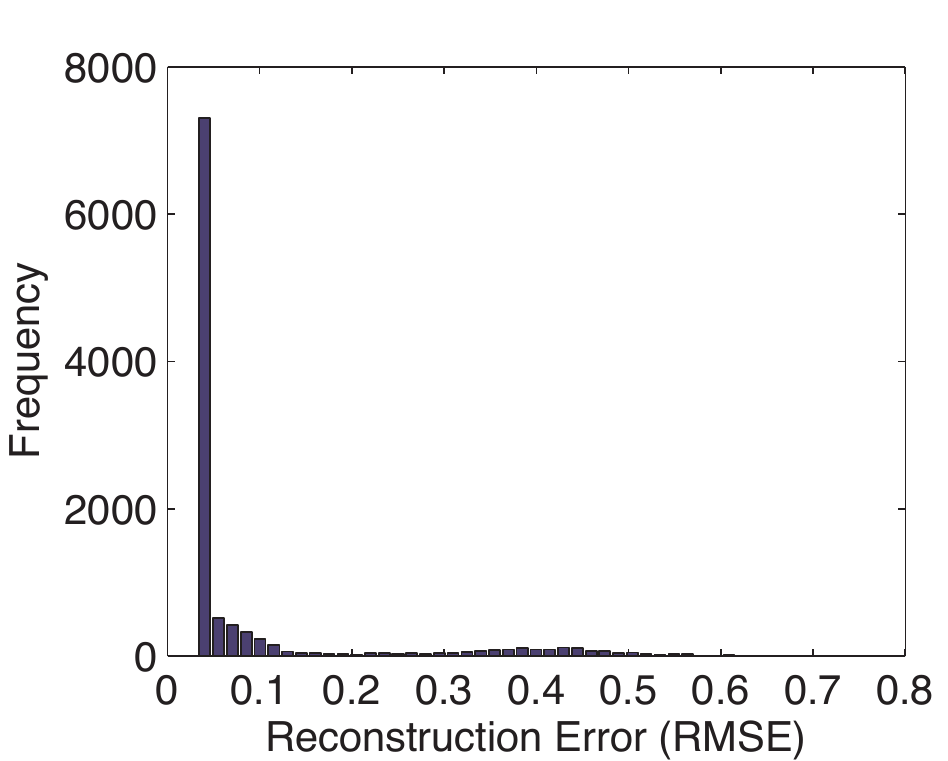}\label{fig:errorDist}}
	\subfigure[]{
	\label{fig:recon:variation}
	\includegraphics[width=0.39\linewidth]{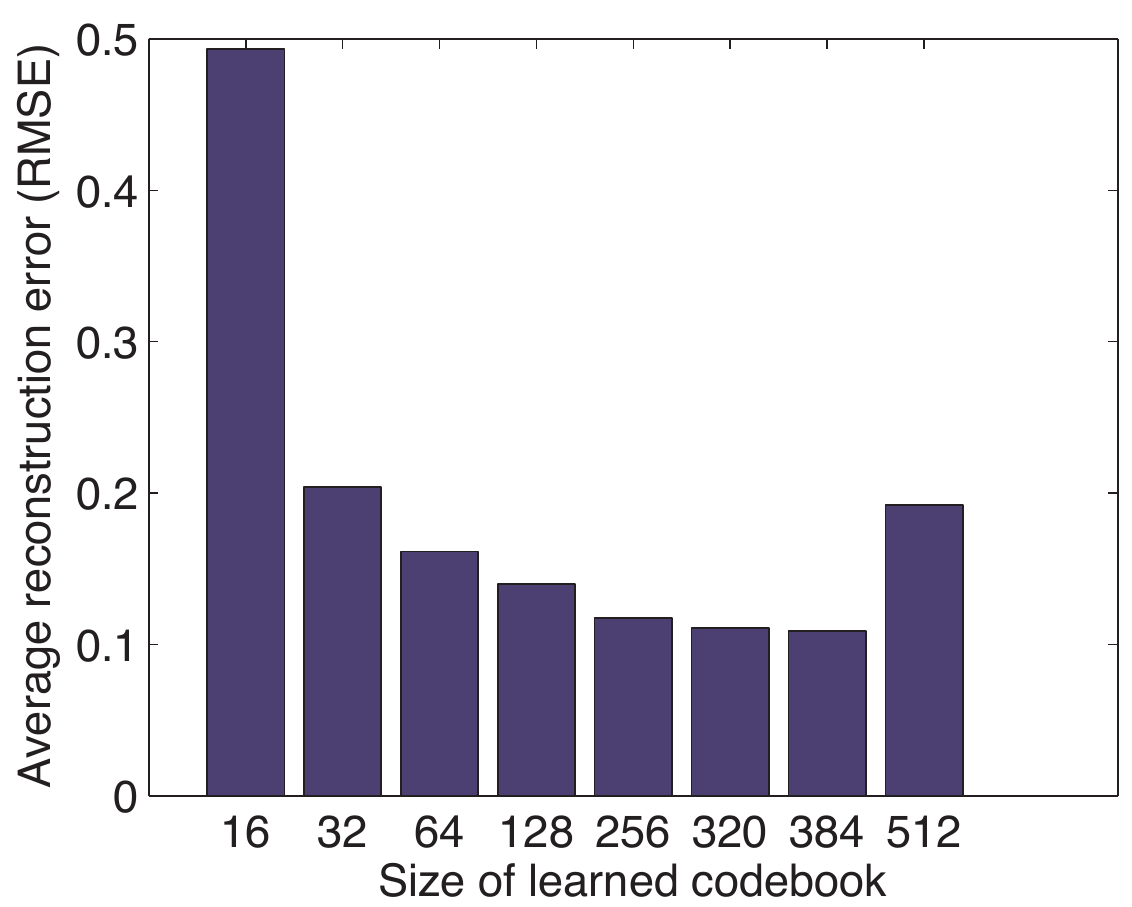}\label{fig:codebookSize}}
\caption{(a) Histogram of the reconstruction error (RMSE) of accelerometer data collected by User 1 using a codebook of $512$ basis vectors. (b) Variation of average reconstruction error with varying codebook size.}
\label{fig:recon}
\end{center}
\end{figure}

The quality of the codebook can be further assessed by computing the average reconstruction error on the unlabeled dataset. 
Figure~\ref{fig:errorDist} shows the histogram of the reconstruction error computed using a codebook of $512$ basis vectors for the dataset collected by User 1. 
The figure indicates that the learned codebook can represent the unlabeled data very well with most of the reconstructions resulting in a small error. 

The reconstruction error on the unlabeled data can also be used to determine the size of the codebook to use. 
To illustrate this, Figure~\ref{fig:codebookSize} shows the average reconstruction error while learning codebooks of varying size from the data collected by User 2. 
Note that the reconstruction error does not necessarily decrease with increased size of the codebook since large over-complete bases can be difficult to learn. 
The figure shows that the codebook with $512$ basis vectors failed to reduce the average reconstruction error, compared to the codebook with $256$ basis vectors. In order to find a good codebook size, we next use a greedy binary search strategy and learn a codebook whose size is halfway in between $256$ and $512$, i.e., $384$. If the new codebook achieves the lowest average reconstruction error, we stop the back tracking (as in this case). Otherwise, we continue searching for a codebook size by taking the mid point between the codebook size with lowest reconstruction error found so far (e.g., $256$) and the latest codebook size tried (e.g., $384$). Figure~\ref{fig:codebookSize} also indicates that an over-complete codebook (i.e., $S \ge 100$), generally improves the accuracy of the data reconstruction.

\begin{figure*}
\begin{center}
\includegraphics[width=0.975\linewidth]{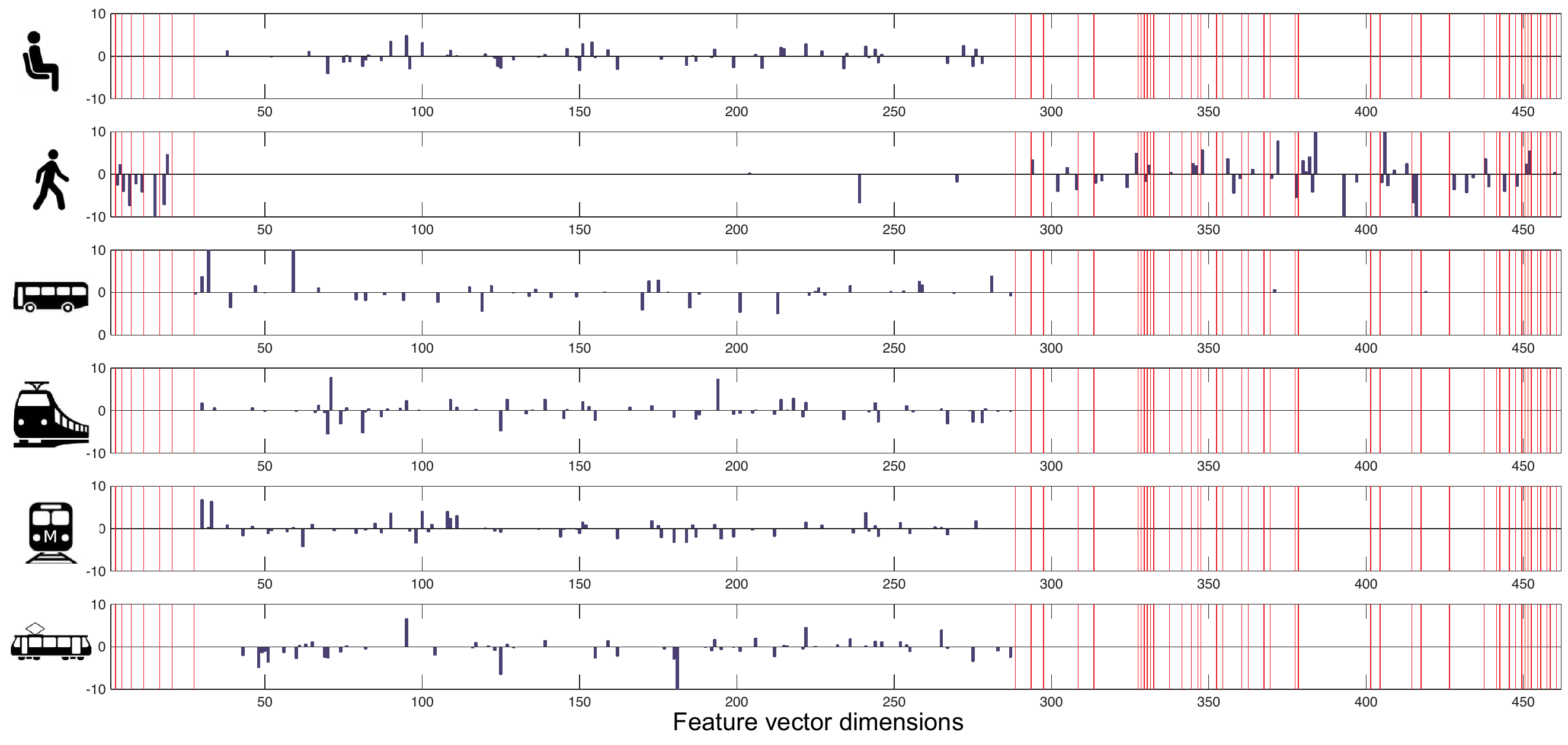}
\caption{Examples of feature vectors derived for different transportation modes using the optimized codebook $\bm{\mathcal{B}}^*$. Vertical lines separate different clusters of basis vectors, which remained after the codebook selection process.}
\label{fig:features}
\end{center}
\end{figure*}

\subsection{Feature Extraction}
\label{codebook}

Examples of features extracted from accelerometer readings collected during different modes of transportation using the optimized codebook $\bm{\mathcal{B}}^*$ are given in Figure~\ref{fig:features}. 
In the figure we have separated activations of basis vectors in different clusters with red vertical lines and the basis vectors within a cluster are sorted based on their empirical entropy. 
Among all transportation modes, `walking', which represents the only kinematic activity in our dataset, is found to be totally different from other activities. 
The figure indicates the presence of a large cluster of basis vectors with structural similarities, which can also be observed from Figure~\ref{fig:dendrogram}. The basis vectors belonging to the large cluster are responsible mostly for capturing inherent patterns present in `static' and `motorized' modes of transportation.

\subsection{Baseline Algorithms}
\label{sec:case-study:baseline}
We compare the effectiveness of the proposed sparse-coding framework with three standard analysis approaches as they have been deployed in a number of state-of-the-art activity recognition applications. In the following, we summarize the technical details of the latter. Note that the focus of our work is on {\em feature extraction}. The classification backend is \textit{principally} the same for all experiments (see Section \ref{sec:results}).

\subsubsection{Principal Component Analysis}
\label{sssec:pca}

Principal Component Analysis (PCA, \cite{Joliffe1986-PCA}) is a popular dimensionality reduction method, which has also been used for feature extraction in activity recognition community \cite{plotz11feature}. 
We use PCA based feature learning as a baseline in our evaluation experiments, and in this section we outline the main differences of PCA compared to the sparse-coding based approach. 

PCA projects data onto an orthogonal lower dimensional linear space such that the variance of the projected data is maximized.
The optimization criterion for extracting principal components, i.e., the basis vectors can be written as:
\begin{eqnarray}
\min_{\bm{\mathcal{B}},\bm{a}} \sum_{i=1}^K || \bm{x}_i - \sum_{j=1}^d a_j^i \bm{\beta}_j||_2^2,\label{eq:pca}\\
\mbox{subject to} \;\;\bm{\beta}_j\perp \bm{\beta}_k, \forall j,k \:\mbox{s.t.} \:\:j \ne k \nonumber
\end{eqnarray} 
where $d$ is the dimensionality of the subspace. Feature vectors $\bm{a}^i$ can be derived by projecting the input data $\bm{x}_i \in \mathcal{R}^n$ on the principal components $\{\bm{\beta}_j\}_{j=1}^d$, where $d \le n$.

PCA has two main differences to sparse-coding. First, PCA extracts only linear features, i.e., the extracted features $\bm{a}^i$ are a linear combination of the input data. 
This results in the inability of PCA to extract non-linear features and restricts its capability for accurately capturing complex activities. 
Second, PCA constraints the basis vectors to be mutually orthogonal, which restricts the maximum number of features that can be extracted to the dimensionality of the input data, i.e., in our case to the frame-length $n$. Hence, PCA cannot extract over-complete and sparse features.

We follow the argumentation in \cite{plotz11feature} and normalize the accelerometer data before applying PCA using an (inverse) ECDF approach. 
The inverse of the empirical cumulative distribution function (ECDF) is estimated for training frames at fixed numbers of points. 
These frame representations are then projected onto the subspace retaining at least $99\%$ of the variance resulting in the final feature representation that is then fed into classifier training.
During the inference, frames from the test dataset are projected onto the same principal subspace as estimated during the training.

\begin{figure}[!t]
\begin{center}
\includegraphics[width=\linewidth]{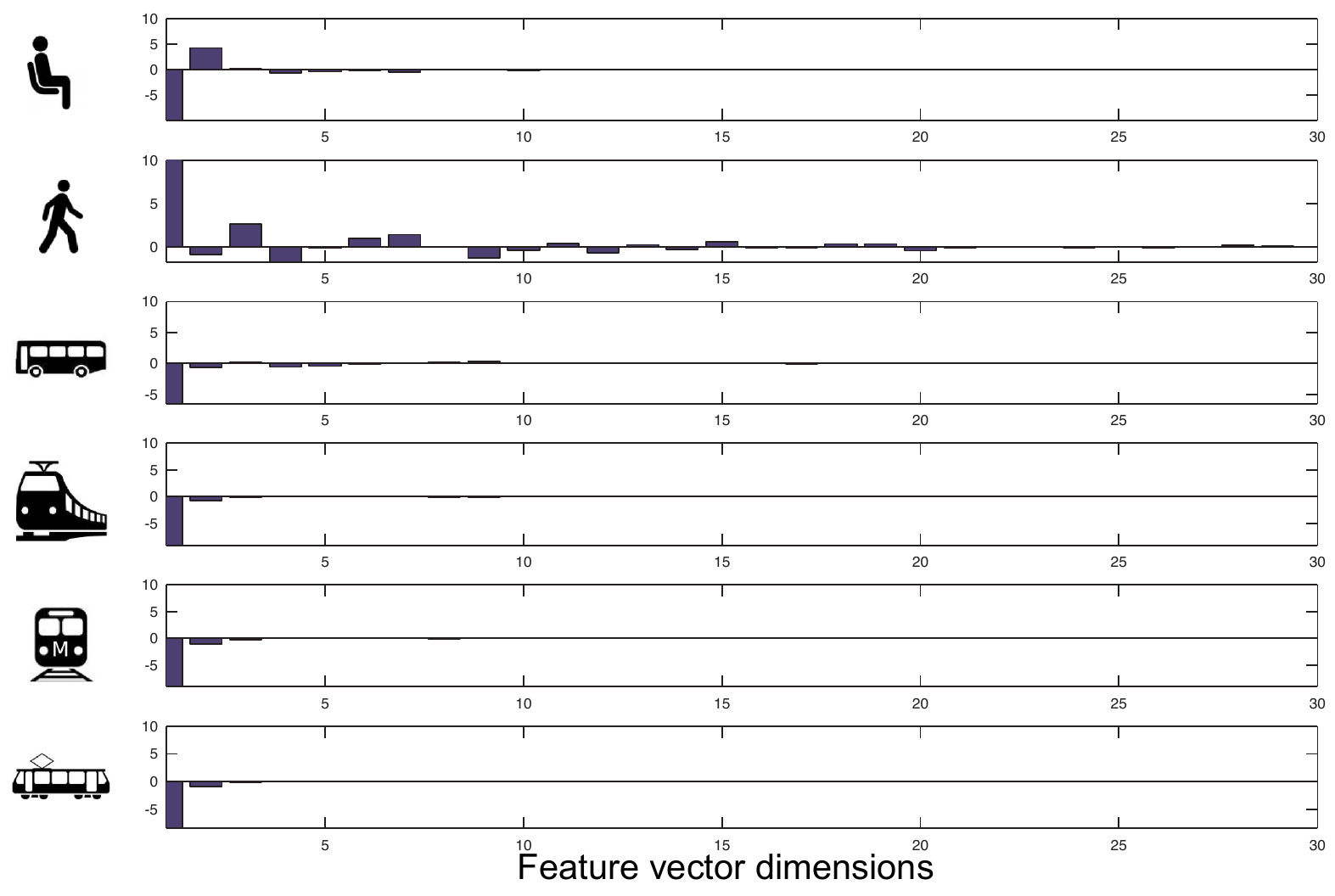}
\caption{Example of  feature vectors obtained using PCA based approach for different transportation modes.}
\label{fig:pcafeatures}
\end{center}
\end{figure}

Figure~\ref{fig:pcafeatures} illustrates PCA features as they have been extracted from the same transportation mode data frames as it was used for the sparse-coding approach, which makes Figures \ref{fig:features} and \ref{fig:pcafeatures} directly comparable.
Input frames of accelerometer readings are projected onto the linear PCA subspace that retains at least 99\% variance of the data, which in our case results in $d=30$-dimensional data vectors. Figure \ref{fig:pcafeatures} indicates that the PCA features are, in general, non-sparse and measurements collected during different motorized transportations are projected to a similar region in the subspace.

\subsubsection{Feature-Engineering}
Aiming for a performance comparison of the proposed sparse-coding framework with state-of-the-art approaches to activity recognition, our second baseline experiment covers feature-engineering, i.e., manual selections of heuristic features.
For transportation mode analysis Wang \textit{et al.}\ have developed a standard set of features that comprises statistical moments of the considered frames and spectral features, namely FFT frequency components in the range $0\sim4$ Hz~\cite{wang10accelerometer}. The details of the extracted features are summarized in Table~\ref{wangFeatures}. 

\begin{table}[!t]
\centering
\begin{tabular}{|l|}
\hline
1) Mean,\\
2) Variance,\\
3) Mean zero crossing rate,\\
4) Third quartile,\\
5) Sum of frequency components between 0$\sim$2 Hz,\\
6) Standard deviation of frequency components \\between 0$\sim$2 Hz,\\
7) Ratio of frequency components between 0$\sim$2 Hz to all \\frequencies,\\
8) Sum of frequency components between 2$\sim$4 Hz,\\
9) Standard deviation of frequency components between \\2$\sim$4 Hz,\\
10) Ratio of frequency components between 2$\sim$4 Hz to \\all frequencies, and\\
11) spectrum peak position.\\

\hline
\end{tabular}
\caption{Features used for feature-engineering experiments \cite{wang10accelerometer}.}
\label{wangFeatures}
\end{table}

\subsubsection{Semi-supervised Learning}
\label{sec:case-study:baseline:semi}
As our final baseline we consider En-Co-Training, a semi-supervised learning algorithm proposed by Guan \textit{et al.}~\cite{guan07activity}. This algorithm first generates a pool of unlabeled data by randomly sampling measurements from the unlabeled dataset. The algorithm then uses an iterative approach to train three classifiers, a decision tree, a Na{\"i}ve-Bayes classifier, and a $3$-nearest neighbor classifier, using the labeled data. For training these classifiers, we use the same features as with our feature-engineering baseline. Next, the three classifiers are used to predict the labels of the samples that are in the pool. Samples for which all classifiers agree are then added to the labeled dataset and the pool is replenished by sampling new data from the unlabeled dataset. This procedure is repeated for a predefined number of times (see~\cite{guan07activity} for details), and the final predictions can be obtained by employing majority voting on the output of the three classifiers.


\section{Results}
\label{sec:results}

We will now report and discuss the results of the transportation mode case study (described in the previous section), thereby aiming to understand to what extent our approach can effectively alleviate the ground truth annotation problem activity recognition systems for ubiquitous/ pervasive computing typically face.

Serving as performance metric for the recognizers analyzed, we compute the $F_1$-score for individual classes of the test dataset:
\begin{eqnarray}
F_1\mbox{-score} = 2\cdot \frac{precision \cdot recall}{precision + recall},
\end{eqnarray}
where $precision$ and $recall$ are calculated in percentages. 
Moreover, in order to mitigate the non-uniform class distribution in the test dataset, we employ the {\em multi-class} $F_1$-score~\cite{sagha11benchmarking}:
\begin{eqnarray}
F_1^M\mbox{-score} =  \frac{\sum_{i=1}^c w_i \cdot F_1^i\mbox{-score}}{\sum_{i=1}^c w_i},
\end{eqnarray}
where $F_1^i$-score represents the $F_1$-score of the $i^{th}$ class (out of $c$ different classes of test dataset) and $w_i$ corresponds to the number of samples belonging to the $i^{th}$ class.

\subsection{Classification Performance}
\label{classPerformance}

\begin{table*}
\centering
\begin{tabular}{|l|r|r|r|r|r|r|r|}
\hline
	& \multicolumn{6}{|c|}{$F_1$-score} & \multirow{2}{*}{$F_1^M$-score}\\
\cline{1-7}	
 \multicolumn{1}{|c|}{Algorithms} & \multicolumn{1}{c|}{Still} 	& \multicolumn{1}{c|}{Walking} & 	\multicolumn{1}{c|}{Bus} & \multicolumn{1}{c|}{Train} & \multicolumn{1}{c|}{Metro} & \multicolumn{1}{c|}{Tram} & \\
\hline
Sparse-coding (this work)		& 	${\mathbf{90.4}}$	& ${\mathbf{98.6}}$& 	$\mathbf{68.6}$	& 	$\mathbf{26.2}$ & 	 $\mathbf{38.4}$&  $\mathbf{44.5}$& $\mathbf{79.9}$\\
En-Co-Training  	& 	$84.0$& 	$97.8$ & 	$55.1$& 	$2.5$& 	$12.0$&	$13.8$ & $69.6$\\
Feature-engineering (Wang et al.) & 	$81.5$	& 	$96.3$& 	$51.3$& 	$2.5$& 	$10.2$& 	$17.3$& 	$67.9$\\
PCA  		& 	$83.9$& 	$91.0$& 	$39.7$& 	$0.2$& 		$3.7$& 		$6.6$&	$65.5$\\
\hline
\end{tabular}
\caption{Classification performance of sparse-coding and baseline algorithms using SVM.}
\label{comparison}
\end{table*}


\begin{table*}
\centering
\begin{tabular}{|l|r|r|r|r|r|r|r|r|r|}
\hline
& \multicolumn{5}{c}{\bf Predictions} & & & &\\
\hline
	&  \multicolumn{1}{c|}{Still}  & \multicolumn{1}{c|}{Walking} &   \multicolumn{1}{c|}{Bus}   &  \multicolumn{1}{c|}{Train}  &  \multicolumn{1}{c|}{Metro}  &  \multicolumn{1}{c|}{Tram}   & \multicolumn{1}{c|}{Precision} & \multicolumn{1}{c|}{Recall}  &\multicolumn{1}{c|}{F$_1$-score}\\
\hline	
Still & 	$37,445$  &   $38$    &   $127$   &   $65$    &   $120$   &   $587$   & $84.2$  & $97.6$ & $90.4$ \\
Walking &  $2$    & $13,052$  &   $169$   &    $6$    &   $11$    &   $50$    & $98.9$ & $98.2$ & $98.6$\\
Bus & $670$   &   $70$    &  $4,682$   &   $87$    &   $219$   &  $1,068$   & $68.4$ & $68.9$ & $68.6$\\
Train & $1,098$   &   $16$    &   $212$   &   $463$   &   $394$   &   $363$   & $46.6$ & $18.2$ & $26.2$  \\
Metro & $1,662$   &    $8$    &   $415$   &   $296$   &  $1,087$   &   $278$   & $56.6$ & $29.0$ & $38.4$\\
Tram & $3,613$   &    $8$    &  $1,245$   &   $76$    &   $91$    &  $2,955$   & $55.7$ & $37.0$  & $44.5$\\
\hline
\multicolumn{7}{r}{\bf Weighted average:} & \multicolumn{1}{r}{$\mathbf{79.5}$}&  \multicolumn{1}{r}{$\mathbf{82.0}$} & \multicolumn{1}{r}{$\mathbf{79.9}$}\\
\end{tabular}
\caption{Confusion matrix for classification experiments using the sparse-coding framework.}
\label{sc}
\end{table*}


The focus of the first part of our experimental evaluation is on the classification accuracies that can be achieved on real-world recognition tasks using the proposed sparse-coding activity recognition approach and comparing it to the results achieved using state-of-the-art techniques (see Section \ref{sec:case-study:baseline}).
Classification experiments on the transportation mode dataset were carried out by means of a six-fold cross validation procedure. 
Sensor readings from one participants ($\sim 2$ hr) were used as the unlabeled dataset (e.g., for codebook estimation in the sparse-coding based approach, see Section~\ref{codebook}), those from the second participant ($\sim 2$ hr) were used as the labeled dataset for classifier training, and the derived classifier is then tested on the remaining set of recordings as collected by the third participant ($\sim 2$ hr) of our case study. 
This procedure is then repeated six times, thereby considering all possible permutations of assigning recordings to the three aforementioned datasets.
The final results are obtained by aggregating over the six folds. 

For our sparse-coding approach we analyzed the effectiveness of codebooks of different sizes. For practicality and also to put a limit on the redundancy (see Section~\ref{sec:system:codebook}), we set an upper bound on the codebook size to $512$.
Based on the reconstruction quality (evaluated on the unlabeled dataset, see Section~\ref{ssec:cblearning}), we derived participant-specific codebooks. In our experiments, the suitable sizes of the codebooks are found to be $512$, $384$, and $512$ respectively. We then construct the optimized codebooks employing the hierarchical clustering followed by the pruning method (see Section~\ref{sec:system:codebook}).
After codebook learning and optimization, the classification backend is trained using the labeled dataset as mentioned before.
Recognizers based on En-Co-Training and PCA (Section~\ref{sec:case-study:baseline}) are trained analogously. 
To ensure the amount of training data does not have an effect on the results, the feature-engineering baseline is trained using solely the labeled dataset.


We use a SVM classifier with all of the algorithms (except En-Co-Training; see Sec.~\ref{sec:case-study:baseline:semi}) in a one-versus-all setting, i.e., we train one SVM classifier for each transportation mode present in the training data. We consider the common choice of radial basis functions (RBF) ($\exp(-\gamma||x-y||^2_2)$) as the Kernel function of the SVM classifiers, and optimize relevant parameters (cost coefficient $C$ and Kernel width $\gamma$) using a standard grid search procedure on the parameter space with nested two-fold cross validation. 
During the prediction phase, we compute the probabilities $p(y^c| \bm{f})$ of each class $y^c$, given an input feature vector $\bm{f}$. The final prediction is then the class with the highest estimated probability, i.e., $y = \arg \max_c p(y^c|\bm{f})$.

Classification results are reported in Table~\ref{comparison}.
It can be seen that 
the novel sparse-coding based analysis approach achieves the best overall performance with a $F_1^M$-score of $79.9\%$. 
In comparison to the three baseline methods, our sparse-coding framework achieves superior performance with all transportation modes. 
The confusion matrix shown in Table~\ref{sc} provides a more detailed picture of the classification performance of our approach.

Verifying the state-of-the-art in transportation mode detection, all considered approaches achieve good performance on walking and stationary modalities.
However, their classification accuracies substantially drop on more complex modalities, i.e., those exhibiting more intra-class variance such as `motorized' ones like riding a bus. 
In fact, the state-of-the-art approaches to transportation mode detection use GPS, GSM and/or WiFi for aiding the detection of motorized transportation modalities, as it has been shown that these are the most difficult modalities to detect solely based on accelerometer measurements \cite{reddy10using}.

The semi-supervised En-Co-Training algorithm has the second best performance overall, with a $F_1^M$-score of $69.6\%$. The feature-engineering approach of Wang et al.\ achieves the next best performance, with a $F_1^M$-score of $67.9\%$, and the PCA-based approach has the worst performance with a $F_1^M$-score of $65.5\%$. Significance tests, carried out using McNemar $\chi^2$-tests with Yates' correction~\cite{mcnemar47note}, indicate the performance of our sparse-coding approach to be significantly better than the performances of all the baselines ($p < 0.01$). Also the differences between En-Co-Training and Wang et al., and Wang et al.\ and PCA were found statistically significant ($p < 0.01$).



To obtain a strong upper bound on the performance of the feature-engineering based baseline, we ran a separate cross-validation experiment where the corresponding SVM classifier was trained with data from two users and tested on the remaining user. This situation clearly gives an unfair advantage to the approach of Wang et al.\ as it can access twice the amount of training data. With increased availability of labeled training data, the performance of the feature-engineering approach improves to $F_1^M$-score of $74.3\%$ (from $67.9\%$). However, the performance remains below our sparse-coding approach ($79.9\%$), further demonstrating the effectiveness of our approach, despite using significantly smaller amount (half) of labeled data.

\begin{figure}
\centering
\includegraphics[width=\linewidth]{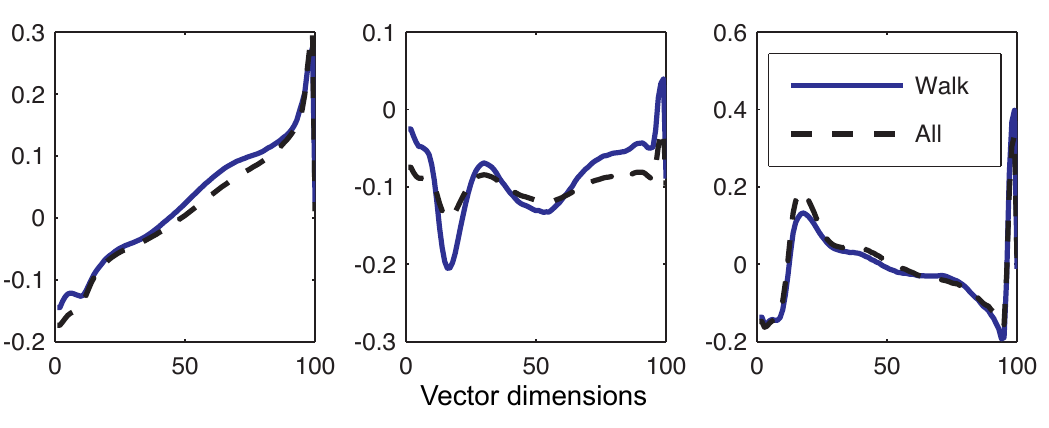}
\caption{First three principal components indicating dominance by a single class with a high variance.}
\label{fig:pcaComponents}
\end{figure}

Analyzing the details of the transportation mode dataset unveils the structural problem PCA-based approaches have.
More than $99\%$ of the frame variance corresponds to the `walking' activity, which results in a severely skewed class distribution.
While this is not unusual for real-world problems it renders pure variance-based techniques --- such as PCA --- virtually useless for this kind of applications.
In our case, the derived PCA feature space captures `walking' very well but disregards the other, more sparse, classes.
The reason for this is that the optimization criterion of PCA aims for maximizing the coverage of the variance of the \textit{data} -- not those of the classes.
In principle, this learning paradigm is similar for any unsupervised approach. 
However, ``blind'' optimization as performed by PCA techniques suffer substantially from skewed class distributions, whereas our sparse-coding framework is able to neutralize such biases to some extent.

In order to illustrate the shortcoming of PCA, Figure~\ref{fig:pcaComponents} illustrates the first three principal components as derived for the transportation mode task. 
Solid blue lines represent the component identified from only `walking' data and the black dashed lines show the case when data from different modes of transportation is used as well for PCA estimation. 
A close structural similarity indicates that the linear features extracted by PCA are highly influenced by one class with high variance, thereby affecting the quality of features for other classes.

\begin{figure*}
\centering
\includegraphics[width=0.8\linewidth]{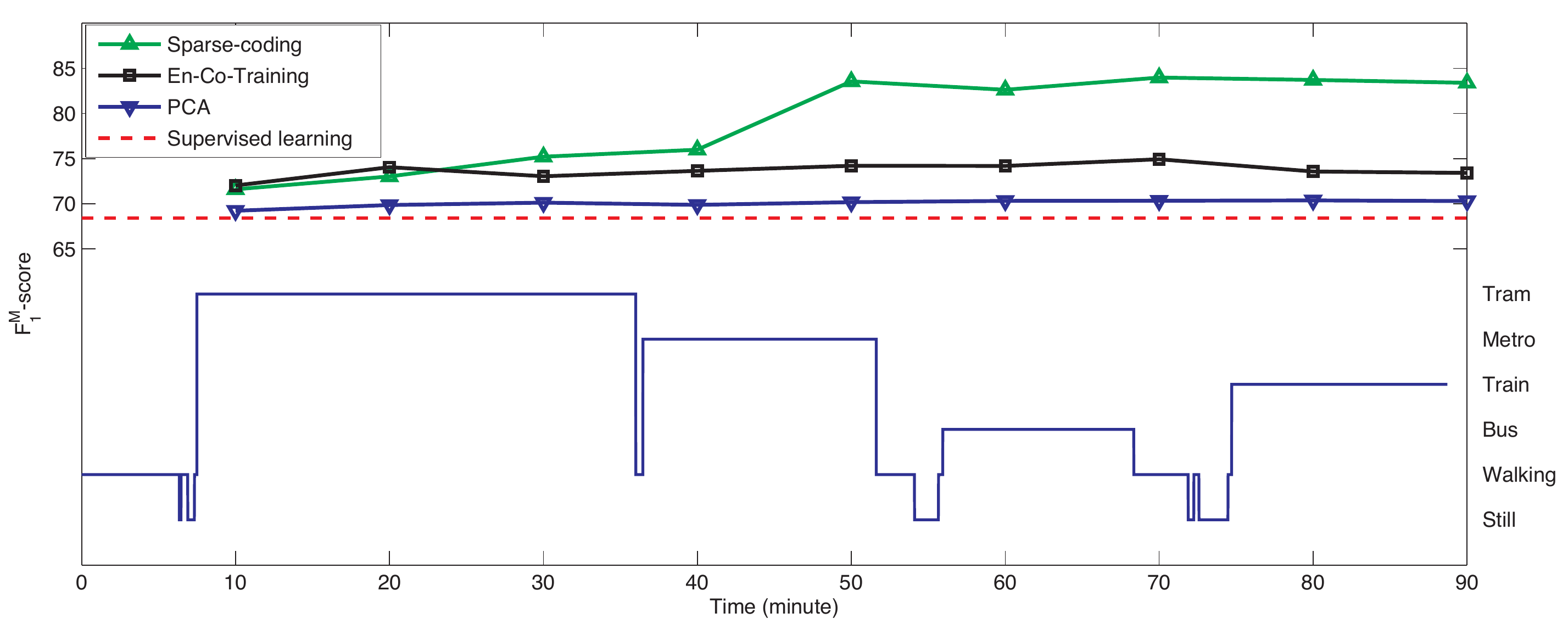}
\caption{Classification accuracies for varying amounts of unlabeled data used (training and test datasets kept fixed).}
\label{fig:semisupervised}
\end{figure*}

For completeness, we have repeated the same six fold cross-validation experiment using C$4.5$ decision trees (see, e.g., \cite[Chap.\ 4]{tan05introduction}) as the classifiers. Similarly to the previous results, the best performance is achieved by the sparse-coding algorithm $75.8\%$. The second best performance, $70.8\%$, is shown by the feature-engineering algorithm of Wang et al., significantly lower than the sparse-coding ($p < 0.01$). The performance of the En-Co-Training remained the same ($69.6\%$) and no significant difference was found compared to the feature-engineering approach. As before, the PCA-based algorithm showed the worst performance ($64.9\%$).



\subsection{Exploiting Unlabeled Data}
\label{ssec:exploit}
The effectiveness of sparse coding depends on the quality of the basis vectors, i.e., how well they capture patterns that accurately characterize the input data. 
One of the main factors influencing the quality of the basis vectors is the amount of unlabeled data that is available for learning. 
As the next step in our evaluation we demonstrate that even small amounts of additional unlabeled data can effectively be exploited to significantly improve activity recognition performance. 

We use an evaluation protocol, where  we keep a small amount of labeled data ($\sim 15$ min) and a test dataset ($\sim 2$ hr) fixed. 
We only increase the size of the unlabeled dataset. 
In this experiment, the training dataset consists of a stratified sample of accelerometer readings from one participant (User 1) only, amounting to roughly $15$ minutes of transportation activities. 
As the test data we use all recordings from User 2. 
We then generate an increasing amount of unlabeled data $\bm{X}(t) $ by taking the first $t$ minutes of accelerometer recordings collected by User 3, where $t$ is varied from $0$ minutes to $90$ minutes with a step of $10$ minutes. 
This procedure corresponds to the envisioned application case where users would carry their mobile sensing device while going about their everyday business, i.e., not worrying about the phone itself, their activities and their annotations. Similarly to the previous section, we use SVM as the classifier.

Figure~\ref{fig:semisupervised} illustrates the results of this experiment and compares sparse-coding against the baseline algorithms. 
In addition to the classification accuracies achieved using the particular methods (upper part of the diagram), the figure also shows the transportation mode ground truth for the additional unlabeled data (lower part). 
Note that this ground truth annotation is for illustration purposes only and we do not use the additional labels for anything else.

We first applied the pure supervised feature-engineering based approach by Wang \textit{et al.}, thereby effectively neglecting all unlabeled data. 
This baseline is represented by the dashed (red) line, which achieves a $F_1^M$-score of $68.2\%$.
Since the supervised approach does not exploit unlabeled data at all, its classification performance remains constant for the whole experiment.
All other methods, including our sparse-coding framework, make use of the additional unlabeled data, which is indicated by actual changes in the classification accuracy depending on the amount of additional unlabeled data used. 
However, only the sparse-coding framework actually benefits from the additional data (see below). The performance improvement by the PCA-based approach over the feature-engineering algorithm is marginal. Whereas, the improvements are significant by the En-Co-Training and our sparse-coding framework.
The more additional data is available, the more drastic this difference becomes for the sparse-coding.

Our sparse-coding framework starts with an $F_1^M$-score of $71.8\%$ when the amount of unlabeled data is small ($t=10$ minutes).
The unlabeled data at $t=10$ minutes only contains measurements from `still', `walking' and `tram' classes and the algorithm is unable to detect the `train' and `metro' activities in the test dataset. 
At $t=30$ minutes, when more measurements from `tram' have become available, the $F_1$-score for that class improves by approximately $18\%$ absolute (not shown in the figure). 
Additionally, sparse-coding begins to successfully detect `train' and `metro' transportation modes due to its good reconstruction property, even though no samples from either of the classes are present in the unlabeled data. 
With a further increase of the amount of unlabeled data, the performance of sparse-coding improves and achieves its maximum of $84.6\%$ at $t=50$ minutes. The performance of sparse-coding remains at this level (saturation), which is significantly better classification performance ($p < 0.01$) compared to all other methods. 
It is worth noting again that this additional training data can easily be collected by simply carrying the mobile sensing device while going about everyday activities and not requiring any manual annotations. 
 
Analyzing the remaining two curves in Figure~\ref{fig:semisupervised} it becomes clear that both En-Co-Training (black curve) and PCA-based recognizers (blue curve) do not outperform our sparse-coding framework.
The plot indicates that these techniques cannot use the additional unlabeled dataset to improve the overall recognition performance. The unlabeled dataset begins with the walking activity (see ground-truth annotations in Figure~\ref{fig:semisupervised}) and the first $10$ minutes of the unlabeled dataset contains a major portion of the high variance activity data. Thus, the principal components learned from the unlabeled data do not change significantly (see Figure~\ref{fig:pcaComponents}) as more and more low variance data from motorized transportations are added. As the labeled training data is kept constant, the feature set remains almost invariant, which explains almost constant performance shown by the PCA-based feature learning approach when tested on a fixed dataset. 

The En-Co-Training algorithm employs the same feature representation as the supervised learning approach and uses a random selection procedure from the unlabeled data to generate a set on which the classification ensemble is applied. The random selection process suffers from over representation by the large transportation activity as it completely ignores the underlying class distribution. The bias toward the large classes limits the En-Co-Train algorithm to utilize the entire available unlabeled dataset well, especially for activities that are performed sporadically. To mitigate the effect of the random selection, we repeat the En-Co-Train algorithm five times and only report the average performance for different values of $t$. Apart from noise effects no significant changes in classification accuracy can be seen and the classification accuracies remain almost constant throughout the complete experiment.

\subsection{Influence of Training Data Size}
\label{ssec:trainVar}

As the next step in our evaluation, we show that the sparse-coding alleviates the ground-truth collection problem and achieves a superior recognition performance even when a small amount of ground-truth data is available.

To study the influence of the amount of labeled training data on the overall recognition performance, we conduct an experiment where we systematically vary the amount of labeled data and measure the recognition performances of all algorithms using SVM, while keeping the unlabeled and the test datasets fixed. More specifically, $24$ training datasets with increasing size are constructed from the data collected by User 1 by selecting the first (chronological) $p\%$ of all the transportation modes (stratification), where $p$ is varied from $1$ to $4$ with unit step and then $5$ to $100$, with a step of $5$. This approach also suits well for the practical use-case, where a small amount of target class specific training data is collected to train an activity recognizer. As the test dataset we use the entire data of User 2 and use the data of User 3 as the unlabeled dataset. Note that the experimental setting differs from that in Section 5.1 and hence the results of these sections are not directly comparable.

\begin{figure}[!t]
\begin{center}
\includegraphics[width=\linewidth]{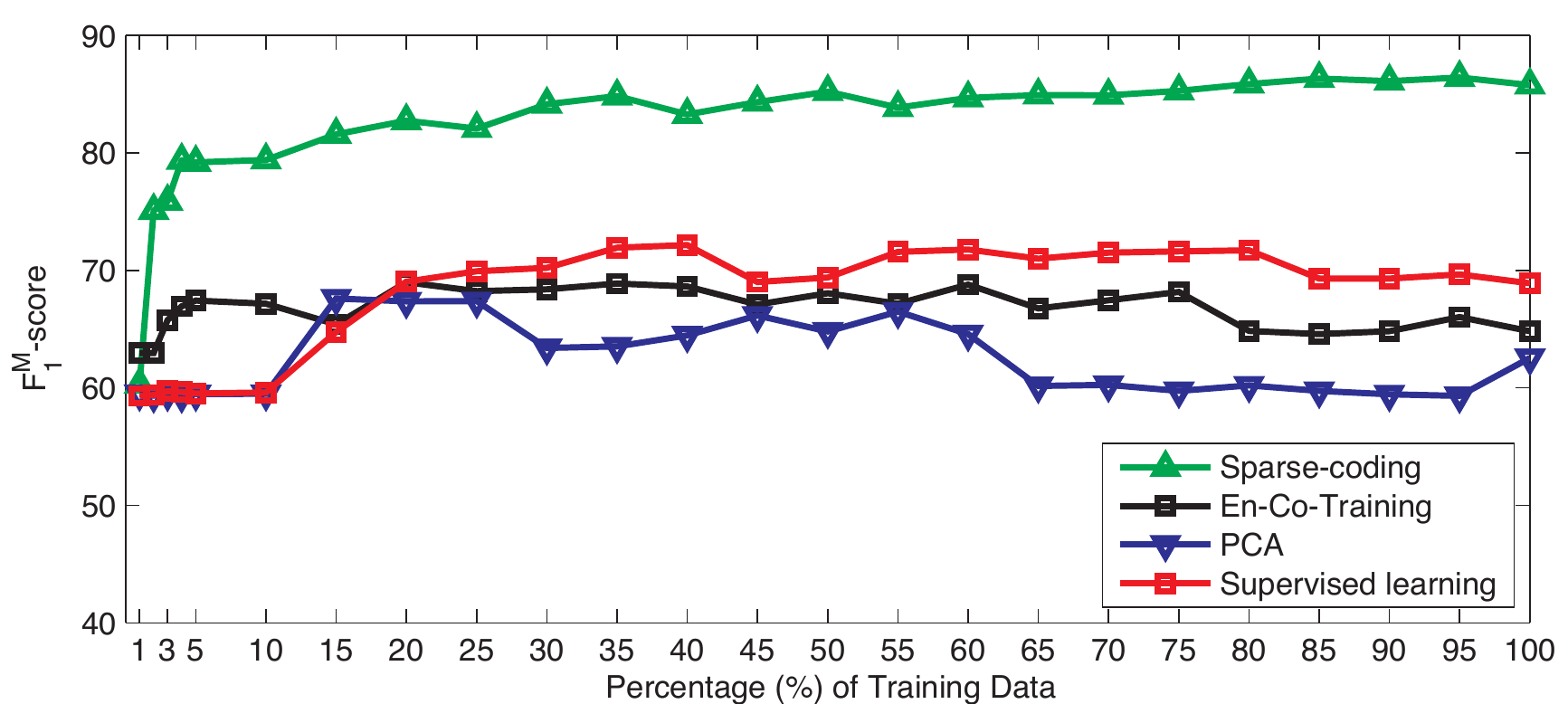}
\caption{Classification accuracies for varying amounts of labeled data used (unlabeled and test datasets kept fixed).}
\label{fig:tariningAmount}
\end{center}
\end{figure}

\begin{table*}[!t]
\centering
\begin{tabular}{|l|r|r|r|r|r|r|r|r|r|}
\hline
	& \multicolumn{8}{|c|}{$F_1$-score} & \multirow{2}{*}{$F_1^M$-score}\\
\cline{1-9}	
 	 \multicolumn{1}{|c|}{Algorithms} &  \multicolumn{1}{c|}{Still} 	&  \multicolumn{1}{c|}{Walking} & 	 \multicolumn{1}{c|}{Bus} &  \multicolumn{1}{c|}{Train} &  \multicolumn{1}{c|}{Metro} &  \multicolumn{1}{c|}{Tram} &  \multicolumn{1}{c|}{{\bf Run}} &  \multicolumn{1}{c|}{{\bf Bike}} & \\
 \hline
Sparse-coding & $\mathbf{88.9}$ & $\mathbf{91.2}$ & $\mathbf{63.8}$ & $\mathbf{24.2}$ & $\mathbf{37.0}$ & $\mathbf{40.8}$ & $95.4$ & $\mathbf{78.8}$ & $\mathbf{79.3}$\\
Feature-engineering (Wang) & $82.3$ & $90.9$ & $61.1$ & $5.2$ & $8.2$ & $17.1$ & $\mathbf{97.9}$ & $68.7$ & $72.2$\\
En-Co-Training & $85.3$ & $89.1$ & $47.3$ & $2.1$ & $12.7$ & $12.9$ & $97.6$ & $56.6$ & $71.2$\\
PCA & $85.0$ & $90.6$ & $39.3$ & $0.7$ & $12.5$ & $11.3$ & $96.6$ & $64.0$ & $70.7$\\
\hline
\end{tabular}
\caption{Classification performance in presence of extraneous activities (`run' and `bike') in the test dataset.}
\label{comparison2}
\end{table*}

\begin{table*}[!t]
\centering
\begin{tabular}{|l|r|r|r|r|r|r|r|r|r|r|r|}
\hline
& \multicolumn{7}{c}{\bf Predictions} & & & &\\
\hline
	&  \multicolumn{1}{c|}{Still}  & \multicolumn{1}{c|}{Walking} &   \multicolumn{1}{c|}{Bus}   &  \multicolumn{1}{c|}{Train}  &  \multicolumn{1}{c|}{Metro}  &  \multicolumn{1}{c|}{Tram}  & \multicolumn{1}{c|}{\bf Run} & \multicolumn{1}{c|}{\bf Bike} & \multicolumn{1}{c|}{Precision} & \multicolumn{1}{c|}{Recall}  &\multicolumn{1}{c|}{F$_1$-score}\\
\hline	
Still &  $37,480$  &   $34$     &  $97$    &   $94$    &   $103$   &   $550$   &  $15$    &    $9$ & $81.6$ & $97.6$ & $88.9$\\
Walking &   $19$    &  $12,293$  &     $221$   &   $25$    &   $10$    &   $58$    & $439$   &   $225$   & $90.0$ & $92.5$ & $91.2$\\
Bus &   $1,240$   &   $24$    &  $4,242$   &   $118$   &   $278$   &   $814$   & $42$    &   $38$    & $65.3$ & $62.4$ & $63.8$ \\
Train &   $1,139$   &   $24$    &    $231$   &   $436$   &   $393$   &   $316$   &   $0$    &    $7$    & $41.1$ & $17.1$ & $24.2$\\ 
Metro &  $1,808$   &   $2$    &    $387$   &   $282$   &  $1,051$   &   $211$   &   $0$    &    $5$    & $54.5$ & $28.1$ & $37.0$\\
Tram &  $4,241$   &   $10$    &   $912$   &   $86$    &   $81$    &  $2,589$   &   $64$    &    $5$    & $55.1$  & $32.4$ & $40.8$\\
{\bf Run} &   $0$    &   $401$   &    $11$    &    $4$    &    $0$    &    $6$    & $11,446$  &   $18$    & $94.5$ & $96.3$ & $95.4$\\
{\bf Bike} &   $31$    &   $870$   &   $393$   &   $17$    &   $13$    &   $152$   &   $104$   &  $3,508$   & $92.0$ & $68.9$ & $78.8$\\
\hline
\multicolumn{9}{r}{\bf Weighted average:} & \multicolumn{1}{r}{$\mathbf{79.3}$}&  \multicolumn{1}{r}{$\mathbf{81.4}$} & \multicolumn{1}{r}{$\mathbf{79.3}$}\\
\end{tabular}
\caption{Confusion matrix for classification experiments using the sparse-coding framework in presence of extraneous activities (`run' and `bike').}
\label{scExtra}
\end{table*}

The results of the experiment are shown in Figure~\ref{fig:tariningAmount}. Our sparse-coding based approach clearly outperforms all baseline algorithms, achieving the best $F_1^M$-score for all training data sizes from $p\ge 2\%$ onwards. When using only $2\%$ of the training data ($\sim 3$ minutes), the sparse-coding achieves a $F_1^M$-score of $75.1\%$, which is significantly better ($p < 0.01$) than all other algorithms, irrespective of the amount of training data they used. This indicates superior feature learning capability of the proposed sparse-coding based activity recognition framework. As more training data is provided, the performance of the sparse-coding, in general, improves and the highest $F_1^M$-score of $86.3\%$ is achieved when $95\%$ of the training data is available.

The state-of-the-art supervised learning approach performs poorly when very little training data (e.g., $\le 5\%$) is available and achieves the lowest $F_1^M$-score of $59.5\%$. Additional ground-truth data, e.g., till $40\%$, continue to improve the performance of the algorithm to $72.2\%$. Further increases in training data, however, fail to improve the performance, with the final performance dipping slightly to a $F_1^M$-score of around $70\%$.

Similarly to our approach, the semi-supervised En-Co-Training algorithm is capable of utilizing unlabeled data, achieving a $F_1^M$ score of $67.4\%$ when $5\%$ of the training data is available. With $5-10\%$ training data, En-Co-Training achieves significantly better performance than the feature-engineering and the PCA-based approaches ($p < 0.01$). The performance of the algorithm slightly improves with additional training data, staying at a level of $69\%$ until $40\%$ of training data is used. Further increases to the amount of training data start to make the algorithm sensitive to small scale fluctuations in measurements, causing a slight dip in performance ($66\%$). These fluctuations are due to the inherent random selection process employed by the algorithm (see Sec.~\ref{sec:case-study:baseline:semi}). When more training data becomes available, the feature-engineering approach surpasses the performance of the En-Co-Training despite relying on the same feature set. Despite the ability of En-Co-Training to utilize unlabeled data, its performance remains below sparse-coding, and also below the feature-engineering for larger training set sizes, indicating poor generalization capability for the ensemble learning employed by the algorithm. 


The PCA-based feature learning approach shows a low recognition performance of $59.5\%$, when training data is smallest. With additional training data (e.g., $20\%$) the algorithm shows improved performance ($67.6\%$), however, the improvement diminishes as more training data is added. As described before, the PCA-based approach learns a set of principal components based on the variance present in the dataset, without considering the class information. When a large amount of training data is provided, the orientation of the principal components, biased by the `walking' activity, generates similar features among the kinematic and motorized activities and makes the classification task difficult, resulting in a drop in overall performance.


\subsection{Coping with Extraneous Activities}

In the final part of the transportation mode case study, we now focus on a more detailed analysis of how the recognition systems cope with sensor data that were recorded during extraneous activities, i.e., those that were not originally targeted by the recognition system. While going about everyday business, such extraneous activities can occur and any activity recognition approach needs to cope with such ``open'' test sets.


We study the reconstruction errors as they occur when accelerometer data from extraneous activities are derived using codebooks that previously had been learned in absence of these extraneous activities.
For this evaluation, we collected additional data from `running' and `biking' activities and then extracted features using the sparse-coding framework as described in Section~\ref{codebook}.

\begin{figure}[!t]
\begin{center}
\includegraphics[width=\linewidth]{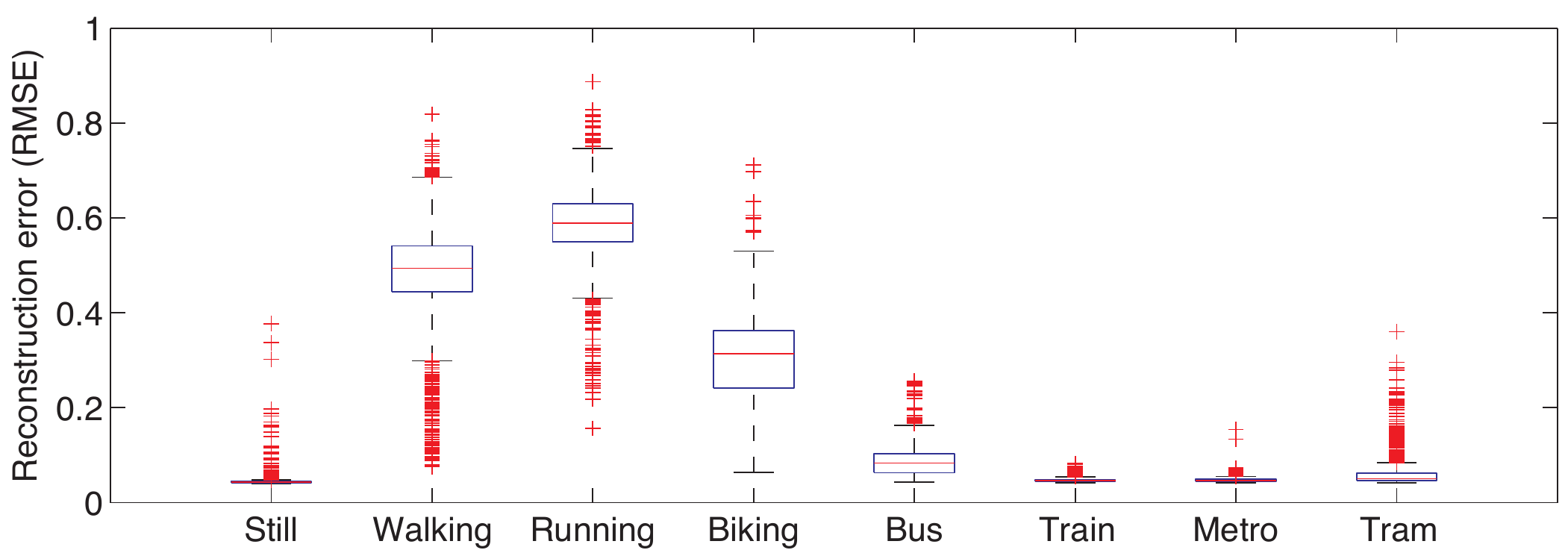}
\caption{Box plot of the reconstruction errors for codebook evaluation on test dataset including previously unseen activities (`running' and `biking').}
\label{fig:boxErr}
\end{center}
\end{figure}

Figure~\ref{fig:boxErr} shows the box-and-whisker diagram, highlighting the {\em quartiles} and the {\em outliers} of the reconstruction errors, for different activities using the same codebook that was used in the experiments of Section~\ref{classPerformance}. 
It can be seen that the learned codebook effectively generalizes beyond the sample activities seen during training.
Although no sensor readings for `running' or `biking' activities were used for learning the codebook, the derived basis vectors
can effectively be used to represent these unseen activities with a reconstruction error that is comparable to those activities that were present during codebook training.
Note that the differences in reconstruction errors reflect actual differences in measurement variance between the activities~\cite{kjaegaard11energy, bhattacharya14robust}, and that generally the higher the variance in measurements, the higher the reconstruction error as more basis vectors are needed to reconstruct the signal (see Figure~\ref{fig:approx}).

For completeness we also repeated the classification experiments as described in Section~\ref{classPerformance} (six-fold cross validation using SVM). We extended the labeled training set by adding approximately $1$ minute of `running' ($60$ frames) and `biking' data ($60$ frames), and added around $10$ minutes of `biking' and $15$ minutes of `running' data to the test set. The results of this cross validation experiment are summarized in Table~\ref{comparison2}. Even in the presence of novel activities, the sparse-coding based activity recognition approach achieves the highest overall $F_1^M$-score of $79.3\%$, which is significantly better than all other approaches ($p < 0.01$ for all). The second best performance is achieved by the feature-engineering approach ($72.2\%$), followed by the En-Co-Training approach ($71.2\%$). Similarly to the earlier experiments, PCA results in the lowest performance at $70.7\%$.

The addition of more high variance kinematic activities, decreases the accuracy of `walking' (previously the only high variance activity) detection for {\em all} algorithms, as the underlying classifiers confuse `walking' with `running' and `biking' activities. The confusion matrix for the sparse-coding algorithm is given in Table~\ref{scExtra}, which shows that the SVM classifier confuses, mainly among the `walking', `running', and `biking' activities. Additionally, some degree of confusion is also observed among the motorized transportation modes and the extraneous activities. Hence, in case of the sparse-coding algorithm, the $F_1$-scores for all the activities degrade and the overall performance is observed to drop slightly, nevertheless significantly better than all other baseline algorithms. The results given in Table~\ref{comparison2} suggest that the recognition performance of activities using a sub-optimal codebook may suffer as the extraneous activities are not represented in the unlabeled dataset. Note that this issue is unlikely to occur in practice as small amounts of the training data (e.g., 1 minute) could be included as part of the unlabeled data and the codebook could be learned from the expanded unlabeled set. As demonstrated in the previous sections, our approach can effectively generalize even from small amounts of unlabeled data.



\section{Generalization: Sparse Coding for Analysis of \\Activities of Daily Living (Opportunity)}
\label{sec:generalization}

In order to demonstrate the general applicability of the proposed sparse-coding approach beyond the transportation mode analysis domain, 
we now report results on an additional activity recognition dataset that covers domestic activities as they were recorded in the \textit{Opportunity} dataset~\cite{Roggen10collecting, lukowicz10recording}.

Opportunity represents the de-facto standard dataset for activity recognition research in the wearable and ubiquitous computing community. 
It captures human activities within an intelligent environment, thereby combining measurements from $72$ sensors with $10$ different modalities. 
These sensors are: 
\textit{(i)} embedded in the environment; 
\textit{(ii)} placed in objects; and 
\textit{(iii)} attached to the human body to capture complex human activity traits. 
To study the performance of our sparse-coding framework, we use the publicly available {\em challenge dataset}\footnote{\url{http://www.opportunity-project.eu/challengeDataset} [Accessed: \today].} and focus on the task B2, i.e., gesture recognition\footnote{\url{http://www.opportunity-project.eu/node/48\#TASK-B2} [Accessed: \today].}. The task involves identifying gestures performed with the right-arm from unsegmented sensor data streams. For the purpose of gesture recognition, in this paper we only consider the inertial measuring unit (IMU) attached to the right lower arm (RLA) which was configured to record measurements approximately at a rate of $30$ Hz for all the inbuilt sensors (e.g., accelerometer, gyroscope and magnetometer).

\begin{table*}[!t]
\centering
\begin{tabular}{|l|c|c|c|c|}
\hline
	& \multicolumn{3}{|c|}{$F_1^M$-score} \\
\hline
	& Accelerometer & Gyroscope &  Accelerometer + Gyroscope\\
\hline
Sparse-coding 			& $\mathbf{65.9}$	& $\mathbf{67.2}$	&$\mathbf{66.6}$\\
Feature-Engineering (Pl{\"o}tz et al.) 	& $65.0$			& $66.0$			&$64.9$\\
PCA 					& $63.7$			& $65.3$			&$63.3$\\
\hline
\end{tabular}
\caption{Classification performance of sparse-coding and baseline algorithms on Opportunity dataset.}
\label{oppresult}
\end{table*}

We deploy the sparse-coding based recognition framework as described in the transportation mode case study (Section \ref{sec:evaluation}).
Task specific modifications are minimal and only of technical nature in order to cope with the collected data.
Contrary to the task of transportation mode detection, sensor orientation information is important for separating the different gestures as they have been performed in Opportunity (e.g., opening a door, moving a cup and cleaning table) \cite{Roggen10collecting, lukowicz10recording}. 
Instead of aggregating sensor readings, our sliding window procedure extracts frames by concatenating one second of samples from each axis, i.e., the recordings are of the form:
\begin{eqnarray}
\bm{x}_i &=& \left\{ d^x_{1},\ldots,d^x_{w},d^y_{1},\ldots,d^y_{w},d^z_{1},\ldots,d^z_{w}\right\},
\end{eqnarray}
where $d^x_k, d_k^y$ and $d_k^z$ correspond to the different axes of a sensor in the $k^{th}$ sample within a frame, and $w$ is the length of the sliding window (here $30$). 
Accordingly, each analysis window contains $90$ samples. 
In line with the reference implementation, subsequent frames have an overlap of $50\%$.

In order to systematically evaluate the sparse-coding framework on Opportunity, we first construct the unlabeled dataset by combining the `Drill', `ADL 1', `ADL 2' and `ADL 3' datasets of the three subjects (S1, S2 and S3). 
We also demonstrate the generalizability of our sparse-coding framework to other modalities, i.e., gyroscope. 
Accordingly, we construct the unlabeled datasets from accelerometer and gyroscope measurements and learn sensor specific codebooks comprising of $512$ basis vectors each and then apply the optimization procedure as described in Section~\ref{sec:system:pruning}. 
For performance evaluation we construct the cross-validation dataset by combining `ADL 4' and `ADL 5' datasets of the same three subjects and run a six-fold cross validation using $C4.5$ decision tree classifier.  Table~\ref{oppresult} summarizes the performance of the sparse-coding when features are considered from accelerometer only, gyroscope only and from both the sensors. For comparison we also include the same cross-validation results as obtained using PCA based feature learning with ECDF normalization (Section~\ref{sssec:pca}). 
Additionally we include the performance of a feature-engineering method using the feature set proposed by Pl{\"o}tz \textit{et at.}~\cite{Ploetz2010-ARI}. The feature set captures cross-axial relations and previously has been used successfully on the Opportunity dataset \cite{plotz11feature}.

Table~\ref{oppresult} shows that our sparse-coding framework significantly outperforms the state-of-the-art on the task of analyzing activities of daily living. 
Sparse-coding achieves  $F_1^M$-scores of $65.9\%$, $67.2\%$ and $66.6\%$ respectively while using features from accelerometer, gyroscope and both sensors together.
The feature-engineering approach results in scores of $65.0\%$, $66.0\%$, and $64.9\%$, and the PCA based approach achieves 
 $63.7\%$, $65.3\%$, and $63.3\%$ respectively.  
The McNemar tests prove that improvements by sparse-coding are statistically significant ($p \ll 0.01$, each) for all three sensor configurations.


\section{Practical Considerations}
\label{sec:practical}

When focusing on ubiquitous computing applications (especially using mobile devices), computational requirements play a non-negligible role in system design. Consequently, we now discuss some practical aspects of our sparse-coding framework for activity recognition.

The most time-consuming part of our approach is the construction of the codebook, i.e., the extraction of the basis vectors. The time that is needed for constructing the codebook depends, among other things, on the size of the unlabeled dataset, the number of basis vectors, the sparsity requirement, and the dimensionality of data, i.e., the length of the data windows. The second most time-consuming task is the training of the supervised classifier using the labeled dataset. However, there is no need to perform either of these tasks on the mobile device as {\em online recognition} of activities is possible as long as the codebook and the trained classifier are transferred to the mobile device from remote servers or cloud. 

\begin{figure}[!t]
\begin{center}
\includegraphics[width=0.75\linewidth]{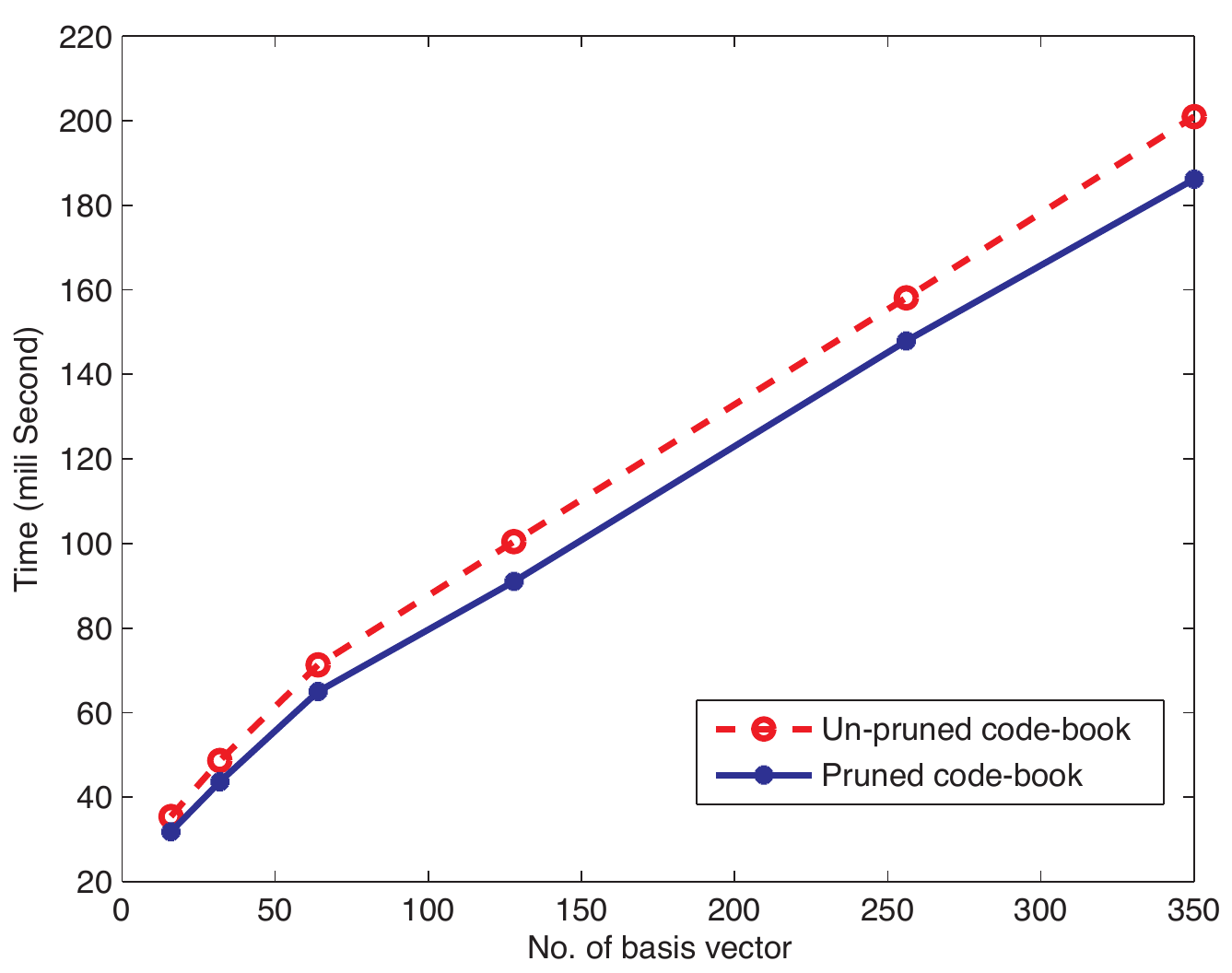}
\caption{Runtime requirements for feature extraction with different codebook sizes, i.e., varying numbers of basis vectors.}
\label{runtime}
\end{center}
\end{figure}

The most computationally intensive task that needs to be performed on the mobile device during online recognition is the mapping of measurements onto the basis vectors, i.e., the optimization task specified by Equation~\ref{feature}. To demonstrate the feasibility of using our framework on mobile devices, we have carried out an experiment where we measured the runtime of the feature extraction using a dataset consisting of $1,000$ frames and with varying codebook sizes. The results of this evaluation are shown in Figure~\ref{runtime}. As expected, the runtime increases as the size of the codebook increases. This increase is linear in the number of basis vectors, with the pruning of basis vectors further reducing the runtime. 
The total time that is needed to run the feature extraction for $1,000$ frames is under $187$ milliseconds (evaluated on a standard desktop PC, solely for the sake of standardized validation experiments) for a codebook consisting of $350$ basis vectors.
With the computational power of contemporary smartphones (such as the Samsung Galaxy SII, which was used for data collection in the transportation mode task) the sparse-coding based approach is feasible for recognition rates of up to $5$ Hz with moderately large codebooks and frame lengths ($1$s). 
This performance is sufficient for typical activity analysis tasks~\cite{bulling14atutorial}.

\section{Summary}
\label{sec:summary}

Ubiquitous computing opens up many possibilities for activity recognition using miniaturized sensing and smart data analysis. 
However, especially for real-world deployments the acquisition of ground truth annotation of activities of interest can be challenging, as activities might be sporadic and not accessible to well controlled, protocol driven studies in a naturalistic and hence representable manner. 
The acquisition of ground truth annotation in these problem settings is resource consuming and therefore often limited. 
This limited access to labeled data renders typical supervised approaches to automatic recognition challenging and often ineffective. 

In contrast, the acquisition of unlabeled data is not limited by such constraints. For example, it is straightforward to equip people with recording devices --- most prominently smartphones --- without the need for them to follow any particular protocol beyond very basic instructions. However, typical heuristic, i.e.\ hand-crafted, approaches to recognition common in this field are unable to exploit this vast pool of data and are therefore inherently limited. 

We have presented a sparse-coding based framework for human activity recognition with specific but not exclusive focus on mobile computing applications.
In a case study on transportation mode analysis we detailed the effectiveness of the proposed approach.
Our sparse-coding technique outperformed state-of-the-art approaches to activity recognition.
We effectively demonstrated that even with limited availability of labeled data, recognition performance of the proposed system massively benefits from unlabeled resources, far beyond its impact on comparable approaches such as PCA.

Furthermore, we demonstrated the generalizability of the proposed approach by evaluating it on a different domain and sensor modalities, namely the analysis of activities of daily living. 
Our approach outperforms the analyzed state-of-the-art in the Opportunity \cite{Roggen10collecting} challenge. 
With a view on mobile computing applications we have shown that --- even if computationally intensive --- inference is feasible on modern, hand-held devices, thus opening this type of approach for mobile applications.

\section{Acknowledgement}
The authors would like to thank Dr.\ P.\ Hoyer for insightful discussions and comments on early versions of this work. The authors also acknowledge Samuli Hemminki for providing help and insights with the transportation mode data.

S.\ Bhattacharya received funding from Future Internet Graduate School (FIGS) and the Foundation of Nokia Corporation.  
Parts of this work have been funded by the RCUK Research Hub on Social Inclusion through the Digital Economy (SiDE; EP/G066019/1), and by a grant from the EPSRC (EP/K004689/1).

\bibliographystyle{../etc/elsarticle-num}

\begin{thebibliography}{10}
\expandafter\ifx\csname url\endcsname\relax
  \def\url#1{\texttt{#1}}\fi
\expandafter\ifx\csname urlprefix\endcsname\relax\def\urlprefix{URL }\fi
\expandafter\ifx\csname href\endcsname\relax
  \def\href#1#2{#2} \def\path#1{#1}\fi

\bibitem{atallah09theuse}
L.~Atallah, G.-Z. Yang, The use of pervasive sensing for behaviour profiling --
  a survey, Pervasive and Ubiquitous Computing 5~(5) (2009) 447--464.
  
\bibitem{bulling14atutorial}
A.~Bulling, U.~Blanke, B.~Schiele, A tutorial on human activity recognition
  using body-worn inertial sensors, ACM Computing Surveys (CSUR) 46~(3) (2014)
  33:1--33:33.

\bibitem{lane10survey}
N.~D. Lane, E.~Miluzzo, H.~Lu, D.~Peebles, T.~Choudhury, A.~T. Campbell, A
  survey of mobile phone sensing, IEEE Communications Magazine 48~(9) (2010)
  140--150.

\bibitem{bao04activity}
L.~Bao, S.~S. Intille, Activity recognition from user-annotated acceleration
  data, in: A.~Ferscha, F.~Mattern (Eds.), Proc. Int. Conf. Pervasive Comp.
  (Pervasive), 2004.

\bibitem{logan07long}
B.~Logan, J.~Healey, M.~Philipose, E.~M. Tapia, S.~Intille, A long-term
  evaluation of sensing modalities for activity recognition, in: Proc. ACM
  Conf. Ubiquitous Comp. (UbiComp), 2007.

\bibitem{pham09slice}
C.~Pham, P.~Olivier, Slice{\&}dice: Recognizing food preparation activities
  using embedded accelerometers, in: Proc. Int. Conf. Ambient Intell. (AmI),
  2009.

\bibitem{Hoey2011-RSA}
J.~Hoey, T.~Pl{\"o}tz, D.~Jackson, A.~Monk, C.~Pham, P.~Olivier, {Rapid
  specification and automated generation of prompting systems to assist people
  with dementia}, Pervasive and Ubiquitous Computing 7~(3) (2011) 299--318.
\newblock \href
  {http://dx.doi.org/http://dx.doi.org/10.1016/j.pmcj.2010.11.007}
  {\path{doi:http://dx.doi.org/10.1016/j.pmcj.2010.11.007}}.
  
  
\bibitem{ploetz12automatic}
T.~Pl\"{o}tz, N.~Y. Hammerla, A.~Rozga, A.~Reavis, N.~Call, G.~D. Abowd, Automatic assessment of problem behavior in individuals with developmental disabilities, in: Proceedings of the 2012 ACM Conference on Ubiquitous Computing.  
  

\bibitem{consolvo08ubifit}
S.~Consolvo, D.~W. McDonald, T.~Toscos, M.~Y. Chen, J.~Froehlich, B.~Harrison,
  P.~Klasnja, A.~LaMarca, L.~LeGrand, R.~Libby, I.~Smith, J.~A. Landay,
  Activity sensing in the wild: a field trial of ubifit garden, in: Proc. ACM
  SIGCHI Conf. on Human Factors in Comp. Systems (CHI), 2008.

\bibitem{Rabbi2011-PAI}
M.~Rabbi, S.~Ali, T.~Choudhury, E.~Berke, Passive and in-situ assessment of
  mental and physical well-being using mobile sensors, in: Proc. ACM Conf.
  Ubiquitous Comp. (UbiComp), 2011.

\bibitem{lester06practical}
J.~Lester, T.~Choudhury, G.~Borriello, A practical approach to recognizing
  physical activities, in: Proc. Int. Conf. Pervasive Comp. (Pervasive), 2006.

\bibitem{parkka06activity}
J.~P{\"a}rkk{\"a}, M.~Ermes, P.~Korpip{\"a}{\"a}, J.~M{\"a}ntyj{\"a}rvi,
  J.~Peltola, I.~Korhonen, Activity classification using realistic data from
  wearable sensors, Biomedicine 10~(1) (2006) 119--128.

\bibitem{bishop07pattern}
C.~M. Bishop, Pattern Recognition and Machine Learning, Springer, 2007.

\bibitem{figo10preprocessing}
D.~Figo, P.~Diniz, D.~Ferreira, J.~Cardoso, Preprocessing techniques for
  context recognition from accelerometer data, Pervasive and Mobile Computing
  14-7 (2010) 645--662.

\bibitem{huynh05analyzing}
T.~Huynh, B.~Schiele, Analyzing features for activity recognition, in: Proc.
  Joint Conf. on Smart objects and Ambient Intell. (sOc-EUSAI), 2005.

\bibitem{kononen10automatic}
V.~K{\"o}n{\"o}nen, J.~M{\"a}ntyj{\"a}rvi, H.~Simil{\"a}, J.~P{\"a}rkk{\"a},
  M.~Ermes, Automatic feature selection for context recognition in mobile
  devices, Pervasive and Ubiquitous Computing 6~(2) (2010) 181--197.

\bibitem{huynh08discovery}
T.~Huynh, M.~Fritz, B.~Schiele, Discovery of activity patterns using topic
  models, in: Proc. ACM Conf. Ubiquitous Comp. (UbiComp), 2008, pp. 10--19.
\newblock \href {http://dx.doi.org/http://doi.acm.org/10.1145/1409635.1409638}
  {\path{doi:http://doi.acm.org/10.1145/1409635.1409638}}.

\bibitem{stikic11weakly}
M.~Stikic, D.~Larlus, S.~Ebert, B.~Schiele, Weakly supervised recognition of
  daily life activities with wearable sensors, IEEE Trans. on Pattern Anal. and
  Machine Intell. (TPAMI) 33~(12) (2011) 2521--2537.

\bibitem{raina07selftaught}
R.~Raina, A.~Battle, H.~Lee, B.~Packer, A.~Y. Ng, Self-taught learning:
  Transfer learning from unlabeled data, in: Proc. Int. Conf. on Machine
  Learning (ICML), 2007.

\bibitem{grosse07shift}
R.~Grosse, R.~Raina, H.~Kwong, A.~Y. Ng, Shift-invariance sparse coding for
  audio classification, in: Proc. Int. Conf. Uncertainty Art. Intell. (UAI),
  2007.

\bibitem{Roggen10collecting}
D.~Roggen, A.~Calatroni, M.~Rossi, T.~Holleczek, K.~F{\"{o}}rster,
  G.~Tr{\"{o}}ster, P.~Lukowicz, D.~Bannach, G.~Pirkl, A.~Ferscha, J.~Doppler,
  C.~Holzmann, M.~Kurz, G.~Holl, R.~Chavarriaga, H.~Sagha, H.~Bayati,
  M.~Creatura, J.~del R.~Mill{\'{a}}n, Collecting complex activity datasets in
  highly rich networked sensor environments, in: Networked Sensing Systems
  (INSS), 2010 Seventh International Conference on, 2010, pp. 233 --240.
\newblock \href {http://dx.doi.org/10.1109/INSS.2010.5573462}
  {\path{doi:10.1109/INSS.2010.5573462}}.

\bibitem{Amft2011-STL}
O.~Amft, {Self-Taught Learning for Activity Spotting in On-body Motion Sensor
  Data}, in: Proc. Int. Symp. Wearable Comp. (ISWC), 2011.

\bibitem{chapelle10semisupervised}
O.~Chapelle, B.~Sch\"olkopf, A.~Zien (Eds.), Semi-Supervised Learning, MIT
  Press, 2010.

\bibitem{guan07activity}
D.~Guan, W.~Y. Lee, Y.-K. Lee, A.~Gavrilov., S.~Lee, Activity recognition based
  on semi-supervised learning, in: Proc. IEEE Int. Conf. on Embedded and
  Real-Time Comp. Systems and Applications (RTCSA), 2007.

\bibitem{stikic08exploring}
M.~Stikic, K.~B. Schiele, Exploring semi-supervised and active learning for
  activity recognition, in: Proc. Int. Symp. Wearable Comp. (ISWC), 2008.

\bibitem{nigam00text}
K.~Nigam, A.~K. McCallum, S.~Thrun, T.~Mitchell, Text classification from
  labeled and unlabeled documents using {EM}, Machine Learning -- Special issue
  on information retrieval 39~(2--3) (2000) 103--134.

\bibitem{stiefmeier08wearable}
T.~Stiefmeier, D.~Roggen, G.~Tr{\"o}ster, G.~Ogris, P.~Lukowicz, Wearable
  activity tracking in car manufacturing, IEEE Pervasive Computing 7~(2) (2008)
  42--50.
  
\bibitem{kjaegaard11energy}
M.~B.~Kj{\ae}rgaard, S.~Bhattacharya, H.~Blunck, P.~Nurmi, Energy-efficient 
   trajectory tracking for mobile devices, in the 9th International Conference on 
   Mobile Systems, Applications and Services, pp. 307-320, 2011.


\bibitem{bhattacharya14robust}
S.~Bhattacharya, H.~Blunck, M.~B.~Kj{\ae}rgaard, P.~Nurmi, Robust and Energy-Efficient Trajectory Tracking for Mobile Devices, 
in: IEEE Transactions on Mobile Computing, 2014.


\bibitem{alemar11using}
H.~Alemar, T.~L.~M. van Kasteren, C.~Ersoy, Using active learning to allow
  activity recognition on a large scale, in: Proc. Int. Joint Conf. Ambient
  Intell. (AmI), Springer, 2011.

\bibitem{Caruana1997-MTL}
R.~Caruana, {Multitask Learning}, Machine Learning 28~(1) (1997) 41--75.

\bibitem{hu11cross-domain}
D.~H. Hu, V.~W. Zheng, Q.~Yang, Cross-domain activity recognition via transfer
  learning, Pervasive and Ubiquitous Computing 7~(3) (2011) 344--358.

\bibitem{kasteren10transferring}
T.~L.~M. van Kasteren, G.~Englebienne, B.~J.~A. Kr{\"o}se, Transferring knowledge
  of activity recognition across sensor networks, in: Proc. Int. Conf.
  Pervasive Comp. (Pervasive), 2010.

\bibitem{lane11enabling}
N.~D. Lane, Y.~Xu, H.~Lu, S.~Hu,T.~Choudhury, A.~T. Campbell, F.~Zhao, Enabling
  large-scale human activity inference on smartphones using community
  similarity networks (csn), in: Proc. ACM Conf. Ubiquitous Comp. (UbiComp),
  2011.

\bibitem{Amar2001-MIL}
R.~A. Amar, D.~R. Dooly, S.~A. Goldman, Q. Zhang, {Multiple-instance learning of
  real-valued data}, in: Proc. Int. Conf. on Machine Learning (ICML), 2001.

\bibitem{stikic09activity}
M.~Stikic, B.~Schiele, Activity recognition from sparsely labeled data using
  multi-instance learning, in: Proc. Int. Symp. on Location and
  Context-Awareness (LoCA), 2009.

\bibitem{coates11analysis}
A.~Coates, H.~Lee, A.~Y. Ng, An analysis of single-layer networks in
  unsupervised feature learning, in: Proc. Int. Conf. Art. Intell. and
  Statistics (AISTAT), 2011.

\bibitem{hinton06fastlearning}
G.~E. Hinton, S.~Osindero, Y.-W. Teh, A fast learning algorithm for deep belief
  nets, Neural Computation 18~(7) (2006) 1527--1554.

\bibitem{mantyjarvi01recognizing}
J.~M{\"a}ntyj{\"a}rvi, J.~Himber, T.~Sepp{\"a}nen, Recognizing human motion
  with multiple acceleration sensors, in: Proc. IEEE Int. Conf. on Systems,
  Man, and Cybernetics (SMC), 2001.

\bibitem{plotz11feature}
T.~Pl{\"o}tz, N.~Y. Hammerla, P.~Olivier, Feature learning for activity
  recognition in ubiquitous computing, in: Proc. Int. Joint Conf. Art. Intell.
  (IJCAI), 2011.
  
\bibitem{plotz11activity}
T.~Pl{\"o}tz, P.~Moynihan, C.~Pham, P.~Olivier, Activity recognition and
  healthier food preparation, in: Activity Recognition in Pervasive Intelligent
  Environments, Vol.~4, Atlantis Press, 2011, pp. 313--329.  

\bibitem{Hammerla2013a}
N.~Hammerla, R.~Kirkham, P.~Andras, T.~Pl{\"o}tz, {On Preserving Statistical
  Characteristics of Accelerometry Data using their Empirical Cumulative
  Distribution}, in: Proc. Int. Symp. Wearable Computing (ISWC), 2013.

\bibitem{minnen06discovering}
D.~Minnen, T.~Starner, I.~Essa, C.~Isbell, Discovering characteristic actions
  from on-body sensor data, in: Proc. Int. Symp. Wearable Comp. (ISWC), 2006.

\bibitem{frank10activity}
J.~Frank, S.~Mannor, D.~Precup, Activity and gait recognition with time-delay
  embeddings, in: Proc. AAAI Conf. Art. Intell. (AAAI), 2010.

\bibitem{hoyer02nonnegative}
P.~O. Hoyer, Non-negative sparse coding, in: Proc. IEEE Workshop on Neural
  Networks for Signal Processing, 2002.

\bibitem{lee07efficient}
H.~Lee, A.~Battle, R.~Raina, A.~Y. Ng, Efficient sparse coding algorithms, in:
  Proc. Int. Conf. Neural Information Proc. Systems (NIPS), 2007.

\bibitem{olshausen97sparse}
B.~A. Olshausen, D.~J. Field, Sparse coding with an overcomplete basis set: a
  strategy employed by v1, Vision Research 37 (1997) 3311--3325.

\bibitem{berkhin06survey}
P.~Berkhin, Survey of clustering data mining techniques, in: J.~Kogan,
  C.~Nicholas, M.~Teboulle (Eds.), Grouping Multidimensional Data, Springer,
  2006, pp. 25--71.
\newblock \href {http://dx.doi.org/10.1007/3-540-28349-8}
  {\path{doi:10.1007/3-540-28349-8}}.

\bibitem{lazer09computational}
D.~Lazer, A.~P.~L. Adamic, S.~Aral, A.-L. Barab{\'a}si, D.~Brewer,
  N.~Christakis, N.~Contractor, J.~Fowler, M.~Gutmann, T.~Jebara, G.~King,
  M.~Macy, D.~R. 2, M.~V. Alstyne, Computational social science, Science
  323~(5915) (2009) 721--723.
\newblock \href {http://dx.doi.org/10.1126/science.1167742}
  {\path{doi:10.1126/science.1167742}}.

\bibitem{zheng11urban}
Y.~Zheng, Y.~Liu, J.~Yuan, X.~Xie, Urban computing with taxicabs, in: Proc. ACM
  Conf. Ubiquitous Comp. (UbiComp), 2011.

\bibitem{soper12human}
D.~Soper, Is human mobility tracking a good idea?, Communications of the ACM
  55~(4) (2012) 35--37.

\bibitem{brezmes09activity}
T.~Brezmes, J.-L. Gorricho, J.~Cotrina, Activity recognition from accelerometer
  data on a mobile phone, in: Workshop Proc. of 10th Int. Work-Conference on
  Artificial Neural Networks (IWANN), 2009.

\bibitem{reddy10using}
S.~Reddy, M.~Mun, J.~Burke, D.~Estrin, M.~Hansen, M.~Srivastava, Using mobile
  phones to determine transportation modes, ACM Trans. on Sensor Networks 6~(2)
  (2010) 13:1--13:27.

\bibitem{wang10accelerometer}
S.~Wang, C.~Chen, J.~Ma, Accelerometer based transportation mode recognition on
  mobile phones, Proc. Asia-Pacific Conf. on Wearable Computing Systems.
  
\bibitem{hemminki13accelerometer}
S.~Hemminki, P.~Nurmi, S.~Tarkoma, Accelerometer-based transportation mode
  detection on smartphones, in: Embedded Networked Sensor Systems (SenSys),
  2013.

\bibitem{lu10jigsaw}
H.~Lu, J.~Yang, Z.~Liu, N.~D. Lane, C.~T., C.~A., The jigsaw continuous sensing
  engine for mobile phone applications, in: Proc. ACM Conf. on Embedded
  Networked Sensor Systems, 2010.

\bibitem{Ploetz2010-ARI}
T.~Pl{\"o}tz, P.~Moynihan, C.~Pham, P.~Olivier, {Activity Recognition and
  Healthier Food Preparation}, in: Activity Recognition in Pervasive
  Intelligent Environments, Atlantis Press, 2010.

\bibitem{Joliffe1986-PCA}
I.~Joliffe, Principal Component Analysis, Springer, 1986.

\bibitem{tan05introduction}
P.-N. Tan, M.~Steinbach, V.~Kumar, Introduction to Data Mining, Addison-Wesley
  Longman Publishing Co., Inc., 2005.

\bibitem{mcnemar47note}
Q.~McNemar, \href{http://dx.doi.org/10.1007/BF02295996}{Note on the sampling
  error of the difference between correlated proportions or percentages},
  Psychometrika 12 (1947) 153--157.
\newline\urlprefix\url{http://dx.doi.org/10.1007/BF02295996}

\bibitem{lukowicz10recording}
P.~Lukowicz, G.~Pirkl, D.~Bannach, F.~Wagner, A.~Calatroni, K.~F{\"o}rster,
  T.~Holleczek, M.~Rossi, D.~Roggen, G.~Tr{\"o}ster, J.~Doppler, C.~Holzmann,
  A.~Riener, A.~Ferscha, R.~Chavarriaga, Recording a complex, multi modal
  activity data set for context recognition, in: ARCS Workshops, 2010, pp.
  161--166.

\bibitem{liao07learning}
L.~Liao, D.~J. Patterson, D.~Fox, H.~Kautz, Learning and inferring transportation routines,
Artificial Intelligence 171 5:6 (2007).

\bibitem{sagha11benchmarking}
H.~Sagha, S.~Digumarti, J.~del R~Millan, R.~Chavarriaga, A.~Calatroni,
  D.~Roggen, and G.~Tr\"oster, Benchmarking classification techniques using the {Opportunity} human
  activity dataset, IEEE International Conference on Systems, Man, and Cybernetics (SMC), 2011.

\end{thebibliography}

\balance

\end{document}